%% file: main.tex
\documentclass{article} % For LaTeX2e
\usepackage{iclr2025_conference,times}

\input{math_commands.tex}

\usepackage{hyperref}
\usepackage{url}
\input{preamble}

\usepackage{placeins}
\usepackage{wrapfig}
\usepackage{subcaption}
\usepackage{xspace}
\usepackage{multirow}
\usepackage{supertabular,capt-of}
\usepackage{booktabs} % For better table aesthetics
\usepackage{caption}  % For controlling caption style
\usepackage{tabularx}
\usepackage{scalerel,graphicx,xparse}
\usepackage{makecell}  % For enhanced cell formatting
\title{Vector-ICL: In-context Learning with Continuous Vector Representations}

\iclrfinalcopy
\usepackage{authblk}
\makeatletter \renewcommand\AB@affilsepx{\quad \protect\Affilfont} \makeatother

\renewcommand\Affilfont{\normalfont\normalsize\raggedright} % Change the font and alignment of affiliations

\author[1]{\textbf{Yufan Zhuang}}
\author[2]{\textbf{Chandan Singh}}
\author[2]{\textbf{Liyuan Liu}}
\author[1]{\textbf{Jingbo Shang}}
\author[2]{\textbf{Jianfeng Gao}}

\affil[1]{UC San Diego}
\affil[2]{Microsoft Research}

\newcommand{\method}{V-ICL\xspace}
\newcommand{\methodlong}{Vector-ICL\xspace}

\newcommand{\encoder}{f_{\textrm{enc}}}
\newcommand{\decoder}{f_{\textrm{llm}}}

\begin{document}

\maketitle

\begin{abstract}
\input{tex/s0_abs}
\end{abstract}

\section{Introduction}
\input{tex/s1_introduction}

\input{tex/figures/pipeline_figure}

\section{Related work}

\paragraph{Empirical results of in-context learning}
ICL has empirically shown strong performance in diverse natural-language processing tasks with very few demonstrations~\citep{brown2020language,openai2023gpt4}.
In modern LLMs with long context windows, ICL has even shown performance improvements as the number of demonstrations grows to hundreds or even thousands, sometimes outperforming finetuning~\citep{agarwal2024many,li2023context,bertsch2024context}.
Empirically, different factors play key roles in ICL.
In smaller LLMs, ground-truth demonstrations are not required for in-context learning, while other factors such as the label space, input text distribution, and overall sequence format play an important role~\citep{min2022rethinking}.
Moreover, these LLMs can sometimes achieve strong performance even when demonstrations are intentionally irrelevant or even pathologically misleading~\citep{webson2022prompt}.
% \textcolor{blue}{
Flan-T5~\citep{flant5} revealed that instruction tuning improves few-shot learning by helping LLMs better utilize in-context examples in an encoder-decoder architecture.%}
% argue against the current practice of prompt engineering, showing that  prompts achieve similar downstream performance to instructively good prompts. 
% ~\citet{shi2023large} investigate the distractibility of LLMs and shows that their performance dramatically drops when irrelevant context is included. 
More recently, \citet{wei2023larger} characterize these behaviors of LLMs with respect to model size, and show that larger language models perform in-context learning differently in the presence of flipped or semantically unrelated labels. 
Orthogonally, different works find ways to improve ICL, e.g. by including explanations~\citep{lampinen2022can},
or chaining ICL calls~\citep{morris2023tree}.
% \paragraph{ICL in different domains}
ICL has shown some success in multimodal models~\citep{wu2024visual,jiang2024many} or when applied to tabular data~\citep{zhao2024probing}. 

\paragraph{Understanding ICL}
Many works have investigated ICL and found that it is able to learn linear models~\citep{akyurek2022learning,zhang2023trained}, discrete functions~\citep{bhattamishra2023understanding}, and more general algorithms~\citep{li2023transformers}.
Some works have explicitly connected ICL in specific settings to implementing optimization steps analogous to gradient descent~\citep{mahankali2023one,von2023transformers,ahn2024transformers} and higher-order optimization methods~\citep{dai2023can,fu2023transformers,giannou2024well,zhang2023trained}.
A complementary direction aims to establish statistical complexity and generalization bounds of in-context learning in transformers~\citep{bai2024transformers,li2023transformers,wies2024learnability,wu2023many}. 
Finally, one recent work suggests that ICL may arise from parallel structures in pretraining data~\citep{chen2024parallel}.

\paragraph{Learning to learn in-context }
In contrast to the emergent ICL capabilities of LLMs, existing works have also studied how to explicitly improve ICL.
~\citet{min2022metaicl} propose MetaICL, a meta-training framework for finetuning pretrained LLMs to perform in-context learning on a large and diverse collection of tasks.
% MetaICL outperforms several baselines including emergent in-context learning and multi-task learning followed by zero-shot transfer.
In the tabular domain, TNP~\citep{nguyen2022transformer} and PFNs~\citep{muller2021transformers} train transformer models to perform in-context prediction for a family of functions, which allows in-context generalization to unseen functions after training.  \citet{zhao2023group} also propose meta-learning transformers to in-context learn group preferences, serving as an in-context learned reward model that adapts to diverse group preferences.
% These works present an interesting set of baselines for our work to examine the in-context learning ability of LLMs.

% \paragraph{Natural language representations in fMRI}

% Using the representations from LLMs to help predict brain responses to natural language has become common among neuroscientists studying language processing in recent years~\citep{jain2018incorporating,wehbe_aligning_2014,schrimpf2021neural,TONEVANEURIPS2019,
% % antonello2021low,
% goldstein_shared_2022}.
% % (see \citep{hale_review} and \citep{jain_computational_2023} for reviews).
% This paradigm of using ``encoding models'' \citep{wu_complete_2006} to better understand how the brain processes language has been applied to help understand the cortical organization of language timescales \citep{JAINNEURIPS2020, chen2023cortical}, examine the relationship between visual and semantic information in the brain \citep{popham2021visual}, and explore to what extent syntax, semantics or discourse drives brain activity \citep{caucheteux_disentangling_2021, kauf2023lexical, reddy_can_2020, pasquiou_information-restricted_2023, aw_training_2022, kumar_reconstructing_2022, oota_joint_2022,tuckute2023driving}.

\input{tex/figures/training_figure}
\section{Method: vector context via embedding projection}
\input{tex/figures/main_results_figure}
\input{tex/s3_methods}

\section{Experimental setup}
\input{tex/s3.5_setup}

\input{tex/tables/merged_table}
\section{Results: unlocking versatile applications across modalities}
\input{tex/s4_results}

\section{Analysis}
\input{tex/s4.5_analysis}

\section{Discussion}
\input{tex/s5_discussion}

\section*{Acknowledgement}
Our work is sponsored in part by NSF CAREER Award 2239440, NSF Proto-OKN Award 2333790, as well as generous gifts from Google, Adobe, and Teradata. Any opinions, findings, and conclusions or recommendations expressed herein are those of the authors and should not be interpreted as necessarily representing the views, either expressed or implied, of the U.S. Government. The U.S. Government is authorized to reproduce and distribute reprints for government purposes not withstanding any copyright annotation hereon.

\bibliography{refs}
\bibliographystyle{iclr2025_conference}

\newpage
\appendix
\input{tex/appendix}

\end{document}

%% file: math_commands.tex
\usepackage{amsmath,amsfonts,bm}

\def\eqref#1{equation~\ref{#1}}
\def\1{\bm{1}}

\DeclareMathAlphabet{\mathsfit}{\encodingdefault}{\sfdefault}{m}{sl}
\SetMathAlphabet{\mathsfit}{bold}{\encodingdefault}{\sfdefault}{bx}{n}

%% file: preamble.tex
\usepackage{graphicx}
\usepackage{float}

\usepackage{placeins} %FloatBarrier
\usepackage{multirow}
\usepackage[capitalize]{cleveref}
\usepackage{array}
\newcolumntype{P}[1]{>{\centering\arraybackslash}p{#1}}

\usepackage{cleveref}

\usepackage{fontawesome}

\usepackage{adjustbox}

\crefname{section}{Sec.}{Secs.}%
\usepackage{scrextend} % for addmargin
\usepackage{siunitx}

\usepackage{makecell}
\newcommand{\capitalizefirst}[1]{\MakeUppercase{\expandafter\@car#1\@nil}\expandafter\@cdr#1\@nil}

\usepackage{changepage}   % for the adjustwidth environment

\usepackage[ampersand]{easylist}
\ListProperties(Hide=100, Hang=true, Progressive=2ex, Style*=$\bullet$ , Style2*=$\circ$)

\usepackage{xspace}

%% file: tex/s0_abs.tex
%%% v0
% We explore whether large language models (LLMs) trained only on text can perform in-context learning (ICL) on continuous vectors from different domains.
% To this end, we align input embeddings from various domains with an LLM's embedding space by learning a lightweight projector (typically a linear transformation) without modifying the LLM's weights.

% We find that pretraining these projectors with general language modeling objectives enables LLMs to perform ICL on the resulting vectors, which we call \methodlong.
% Moreover, finetuning these projectors with task-specific data enhances their effectiveness at enabling ICL, even surpassing few-shot ICL in domains where textual examples can be given as context. 
% We conduct extensive experiments across a variety of domains and tasks, such as text reconstruction, numerical function regression, 
% text classification, summarization, molecule captioning, fMRI decoding \& classification, time-series classification, and graph classification. 
% Our results highlight LLMs' competitive performance in ICL with embedding vectors, revealing previously unknown capabilities of these models.

%%% v1
% \jingbo{I saw in later sections, an effective encoder is required here. Maybe mention it (vaguely) here? otherwise, the projector seems heavy instead of lightweight.} 
% \lucas{good point}
Large language models (LLMs) have shown remarkable in-context learning (ICL) capabilities on textual data. 
We explore whether these capabilities can be extended to continuous vectors from diverse domains, obtained from black-box pretrained encoders. 
By aligning input data with an LLM's embedding space through lightweight projectors, we observe that LLMs can effectively process and learn from these projected vectors, which we term Vector-ICL. 
In particular, we find that pretraining projectors with general language modeling objectives enables Vector-ICL, while task-specific finetuning further enhances performance.
In our experiments across various tasks and modalities, including text reconstruction, numerical function regression, text classification, summarization, molecule captioning, time-series classification, graph classification, and fMRI decoding, Vector-ICL often surpasses both few-shot ICL and domain-specific model or tuning. 
We further conduct analyses and case studies, indicating the potential of LLMs to process vector representations beyond traditional token-based paradigms.
\renewcommand\thefootnote{}\footnotetext{Code is available at: \url{https://github.com/EvanZhuang/vector-icl}.}\renewcommand\thefootnote{\arabic{footnote}}

%% file: tex/s1_introduction.tex
%Large language models (LLMs) naturally think in vector forms, transforming discrete tokens through a series of linear and non-linear operations to generate human-readable outputs. 
%\lucas{instead of saying that llm naturally think in vector forms (its a bit ambiguous), being more specific and saying that although LLMs is trained on discrete natural language tokens, they convert discrete word tokens to continuous vectors, and all non-linear transformations happen in contiguous vector forms, }

In-context learning (ICL) has emerged as a powerful paradigm in large language models (LLMs), 
allowing generalization from limited examples within a given context~\citep{brown2020language,openai2023gpt4}.
By providing demonstrations in the context during inference, ICL allows models to adapt to new tasks and formats without the need for retraining.
% Since LLMs are trained on discrete natural language tokens, ICL is is generally learned and used through natural language.
However, since LLMs are trained on discrete natural language tokens, ICL is generally learned and used through natural language, limiting its applicability to non-textual data.

%\lucas{maybe consider to change the order of these two sentences. }

We explore whether LLMs can perform ICL directly on continuous vectors, a capability that could dramatically expand their applicability. Many data modalities, such as sensor readings, financial time series, or scientific measurements, lack a natural text representation. Moreover, even for text data, information like numbers might be better represented via continuous vectors than tokens.

% We explore whether LLMs can perform ICL directly on continuous vectors that is not in the embedding layer. 
% % We instead explore whether LLMs can directly learn from continuous context in ICL. 
% % they convert discrete text tokens to continuous vectors, and all non-linear transformations happen in vector forms.
% %In this work, we ask: \textit{Can LLMs directly learn from continuous context?} \lucas{this sentence is a bit confusing}
% Enabling LLMs to work with continuous inputs could expand their applicability, since many data modalities lack a natural text representation. 
% Moreover, even for text data, information like number might be better represented via continuous vectors than token sequences. 
% % Moreover, while some tasks can be represented as text, doing so may not always be the most effective approach.
% % \lucas{these two sentences are well written. maybe a good naming for our research / method can be named as `language model handling non-language inputs`?}
% % \yufan{I feel like there are also a big part of our work, that is handling language inputs, but i get your point.}

In our study, we observe that LLMs can indeed understand and process continuous context via embedding projection. 
This technique, which we term Vector-ICL, acts as a bridge between continuous data and the LLM's embedding space.
Simple linear projections are often sufficient, though for cross-modal tasks—such as those involving non-textual data like time-series or graphs, non-linear transformations may be required.
We demonstrate that training the embedding projector using a straightforward next-token prediction objective enables~\methodlong, effectively teaching the LLM to ``read'' continuous vectors.
Moreover, fine-tuning the projector on downstream tasks further enhances the effectiveness of continuous context, outperforming few-shot ICL and domain-specific models or tuning.

Our investigation begins with the task of text reconstruction, where we assess whether LLMs can recover information encoded in text embedding. 
This serves as a proof-of-concept for Vector-ICL, showing that LLMs can indeed extract meaningful information from projected continuous vectors.
We then turn to the more complex challenge of arithmetics. 
Although state-of-the-art LLMs can solve Olympiad mathematical problems~\citep{trinh2024solving,openai2023gpt4}, they struggle with precise number processing due to the limitations inherent in their tokenization schemes. 
Our results demonstrate that \methodlong offers a more effective approach for function approximation, particularly for large numbers that span multiple tokens, potentially opening new avenues for enhancing LLMs' numerical reasoning capabilities.

Finally, we extend our analysis to a broad range of modalities and tasks, including text classification, summarization, molecule captioning, brain fMRI reconstruction and classification, time-series classification, and graph classification. 
Across these diverse domains, LLMs exhibit competitive and often superior performance when employing Vector-ICL, revealing previously untapped capabilities of these models. 
This work highlights the potential of continuous representations in enhancing LLMs' in-context learning capacities, pushing the boundaries of what these models can achieve beyond token-based paradigms.

%% file: tex/figures/pipeline_figure.tex
\begin{figure}[t!]
    \begin{center}
    \includegraphics[width=\linewidth, keepaspectratio]{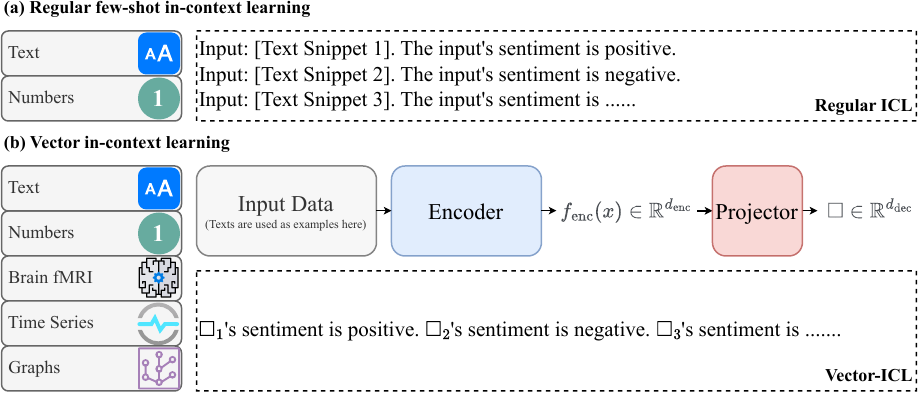}
    \end{center}
    \caption{\textbf{Comparing regular in-context learning to vector in-context learning.}
    % Construct continuous context via embedding projection}
    (a) In regular ICL, textual demonstrations are given as context during LLM inference. 
    (b) In \methodlong, the input space is extended across multiple modalities. 
    %\jingbo{what is the exact requirement of this ``effective''? does the encoder required to be trained as a part of a multi-modal LLM? If so, why do we have to introduce another transformation? If not, why this effective encoder can be transferred to LLM's embedding space?}
    %\yufan{I see, making adjustments}
    The input data is first encoded as embeddings, then transformed into continuous vectors which represent as box tokens ($\Box$) via embedding projection.
    During inference, we provide box tokens in prompts as demonstrations for ICL.
    % into prompts and provide them as continuous context to the LLM. 
    We consider box tokens representing text, numerical data, brain fMRI, time series, and graphs in this study.
    }
    \label{fig:pipeline}
\end{figure}

%% file: tex/figures/training_figure.tex
\begin{figure}[t!]
    \begin{center}
    \includegraphics[width=\linewidth, keepaspectratio]{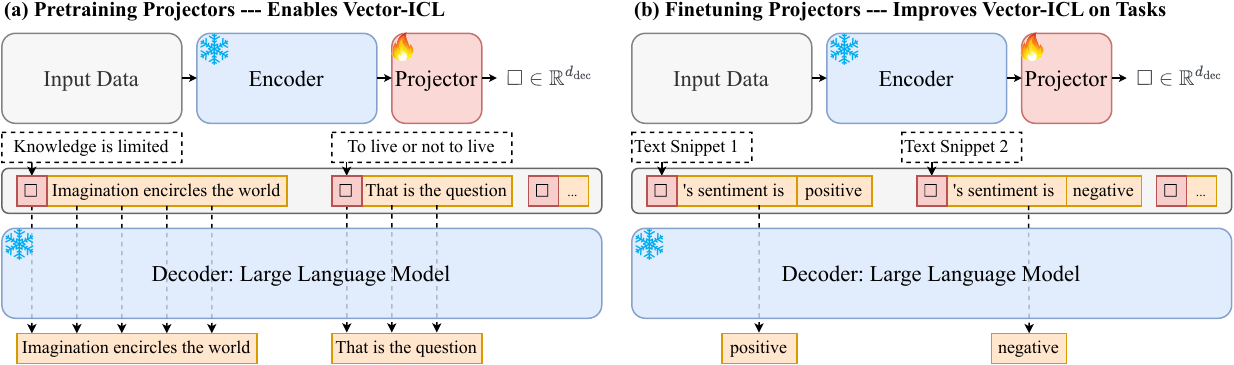}
    \end{center}
    \caption{\textbf{Pretraining and finetuning the projectors.}
    Vector-ICL requires updating the parameters of a lightweight projector while keeping the encoder and decoder parameters fixed.
    The encoder first compresses the input into single token embeddings, and then the projector will project it to the aligned representation space for LLMs' later use.
    %\jingbo{are we always encoding the input data into one token embedding? will there be any limitation caused by this design?}
    %\yufan{added more explanations }
    % only update the projector's parameters.
    (a) Pretraining the projector on a general language modeling corpus (or a modality-to-text dataset) enables \methodlong.
    (b) Task-specific fine-tuning makes \methodlong outperform few-shot ICL on natural language tasks, as well as with domain-specific models on non-language tasks.}
    \label{fig:pipeline_train}
\end{figure}

%% file: tex/figures/main_results_figure.tex
\begin{figure}[t!]
    \begin{center}
    \includegraphics[width=\linewidth, keepaspectratio]{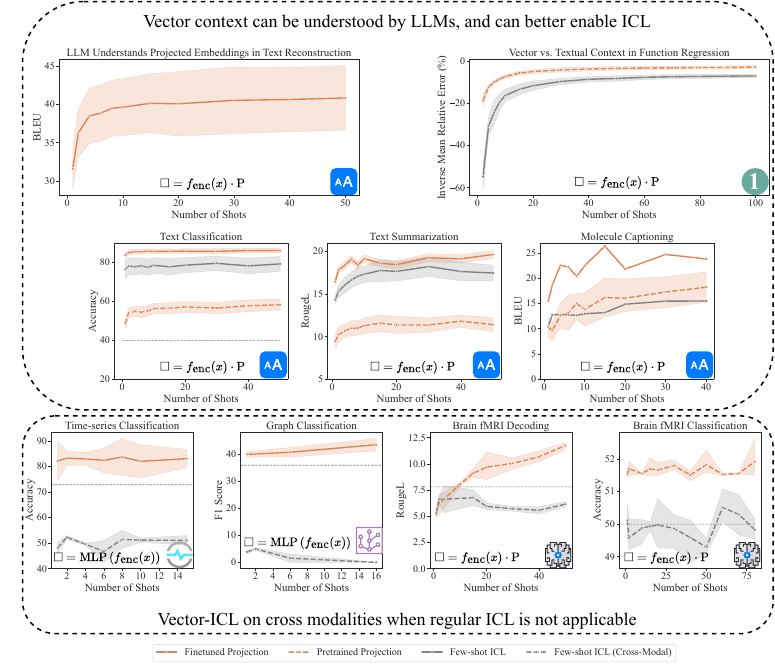}
    \end{center}
    \caption{\textbf{Main results: LLMs can perform \methodlong{} ($\uparrow$ = better).} 
    % This figure presents the aggregated results from our experiments. 
    %%We show that training the embedding projector with a simple next-token prediction objective enables \methodlong. 
    %Fine-tuning the projector on downstream tasks further improves the use of continuous context, surpassing the performance of few-shot ICL and domain-specific baseline models. 
    We show that training the embedding projector with a simple next-token prediction objective enables \methodlong. 
    % \textcolor{blue}{
    Even with only unsupervised pretraining, Vector-ICL matches or outperforms traditional few-shot ICL on 4 out of 6 tasks where direct comparison is possible.
    Fine-tuning the projector on downstream tasks further enhances the use of continuous context, consistently surpassing both few-shot ICL and specialized task-tuned baselines (soft-prompt for text, tuned encoders for non-text).%}
    The study begins with text reconstruction to assess LLMs' ability to interpret box token embeddings, followed by function regression to evaluate reasoning capabilities. 
    We then demonstrate \methodlong's effectiveness and applicability across various downstream tasks, including text classification, summarization, time-series classification, graph classification, and brain fMRI decoding \& classification.
    Results in each panel are averaged over different encoders and LLMs for the diverse tasks we study; error bars show 95\% confidence intervals.
    %\chandan{Minor: you might want to change the red/green to something else for colorblind ppl.}
    }
    \label{fig:main_result}
\end{figure}

%% file: tex/s3_methods.tex
\subsection{Embedding projection}
% We will illustrate how to construct continuous context via embedding projection.
% For any given data $\mathcal{X}=\{x_i\}_{i=1}^n$, we will assume there exists an encoder $\encoder$  for $\mathcal{X}$, or the data is already an abstract representation.
% And we are using a large language model (LLM) as the decoder $\decoder$.
%To perform \methodlong{}, an input must be transformed into a
% We describe the process of constructing
%vector context through an embedding projection. 

\methodlong requires transforming inputs into vector contexts through an embedding projection.
Given a dataset $\mathcal{X} = \{x_i\}_{i=1}^n$, we assume the existence of an encoder $\encoder$, that transforms the data into an abstract representation (alternatively, the raw data may already be a continuous vector).
% or that the data is already provided in vector form. 
The encoded embeddings, $\encoder(x)$, are then projected into box tokens, denoted as $\Box_x$.
Throughout the paper, we will use the terms ``box tokens'' and ``projected embeddings'' interchangeably.
For decoding, we use a language model $\decoder$ that generates outputs based on the provided prompts.
 
% The data can be of any form, we put no limitations on that, we only require the encoder $\encoder$ to map it into a vector space: 
We impose no constraints on the form of the input data $\mathcal{X}$; it can come from any modality. The only requirement is that the encoder $\encoder$ maps each data point $x$ into a vector space, defined as:
\begin{align}
    f_{\mathrm{enc}}: x \rightarrow \mathbb{R}^{d_{\mathrm{enc}}},\; \forall x \in \mathcal{X}
\end{align}
The LLM typically processes discrete tokens $\{\mathrm{tok}_{1}, \mathrm{tok}_{2}, \ldots, \mathrm{tok}_{l}\}$, then maps them to text embedding space $\{\mathrm{emb}_{1}, \mathrm{emb}_{2}, \ldots, \mathrm{emb}_{l}\}, \; \forall i,\;\mathrm{emb}_{i}\in \mathbb{R}^{d_{\mathrm{dec}}}$. 
Since we operate mostly in embedding space, we omit the tokenization step for simplicity and directly refer to text inputs as their embedding representations.

The process of embedding projection is then carried out as follows. 
For linear projection, we construct a projection matrix $\mathrm{P} \in \mathbb{R}^{d_{\mathrm{enc}} \times d_{\mathrm{dec}}}$ and make the following transformations to obtain projected embedding $\Box_x$ given input $x$:
\begin{align}
    \Box_x :=  \encoder(x) \cdot \mathrm{P}
\end{align}
In cases where more expressive power is needed, we utilize a two-layer multi-layer perceptron (MLP) to perform the projection:
\begin{align}
    \Box_x :=  \textrm{MLP}(\encoder(x))
\end{align}
The MLP follows the architecture of the MLP block in Llama~\citep{touvron2023llama2}, with an additional input projection layer to map the input dimension $\mathbb{R}^{d_{\mathrm{enc}}}$ to output dimension $\mathbb{R}^{d_{\mathrm{dec}}}$.

\subsection{Projected embeddings as context}
The projected embeddings are then utilized as context in \methodlong, functioning as the equivalent of the original input data. 
% We incorporate the $\Box$ token into structured prompts as a representation of these projected embeddings.
For example, in NLP tasks, the original text snippets $x$ will first be encoded as embeddings $\encoder{(x)}$, then projected to become $\Box_x$, inserted into the prompt like the following:

\makebox[\textwidth]{$\Box_x$'s sentiment is $\ldots$ / $\Box_x$'s summarization is $\ldots$}

Using them as the context in ICL is then natural:

\makebox[\textwidth]{$\Box_1$'s sentiment is happy.\quad $\Box_2$'s sentiment is sad.\quad $\Box_x$'s sentiment is $\ldots$}

where $\Box_1$ and $\Box_2$ are in-context examples and $\Box_x$ is the input.

\subsection{Training the embedding projectors}
The projectors need to be trained to achieve effective projections. 
We discovered that pretraining these projectors with language modeling objectives enables the ICL capabilities with vector context, and finetuning them on task datasets further improves ICL performance.

The pretraining process is depicted in~\cref{fig:pipeline_train}(a). 
For each text snippet, we cut it into two pieces with the cutting point randomly sampled from the end of sentences.
The first half is encoded and projected while the second is kept intact.
%We use batched grouping, meaning that a single batch contains multiple pairs of projected embeddings and regular text tokens.
The rest is the same with any pretraining process, the language model generated the next token distribution at each input position, except for the ones preceding the projected embeddings, and a cross-entropy loss is imposed on top of this.
With the encoder and LLM frozen, the gradient backpropagates to the projector, updating its parameters.

For non-text data modalities, pretraining can be more flexible. We define this pretraining as involving general, non-task-specific objectives, such as reconstructing a number from its embeddings (e.g., $\Box_x$ is \underline{32768}), performing basic algebra (e.g., $\Box_x$ + $\Box_y$ = \underline{16384}), or predicting the next token from brain fMRI embeddings.

The finetuning process is shown in~\cref{fig:pipeline_train}(b). 
It utilizes additional structured prompts and trains with task-specific datasets. 
Similarly, the input is first mapped into the embedding space and projected into $\Box$ tokens. 
They are then inserted into structured prompts, while the projector is trained with conditional generation loss given those prompts.

%% file: tex/s3.5_setup.tex
\cref{tab:setup} gives an overview of our experimental setup, including specifics for the task, datasets, encoders, LLMs, and task-specific prompts we use.
Across different tasks, we project to four open-weights LLMs.
% We utilize four state-of-the-art LLMs in our experiments: Meta-Llama-3.1-8B-Instruct~\citep{dubey2024llama3},  Mistral-7B-Instruct-v0.3~\citep{jiang2023mistral}, Qwen2-7B-Instruct~\citep{yang2024qwen2}, and Yi-1.5-9B-Chat~\citep{young2024yi}.
We now provide details for individual tasks.
% For each task, we provide comprehensive descriptions, covering the encoders used for embedding construction, datasets, task-specific prompt formulations, and how the embeddings are integrated into the prompts. 
% A summary of these configurations can be found in~\cref{tab:setup}.

\paragraph{Baselines}
% \textcolor{blue}{
We evaluate \methodlong in two distinct settings: with and without task-specific tuning. 
For textual tasks, we compare against few-shot ICL and soft prompt tuning~\citep{li2021softprompt}.
We choose soft prompt tuning as our primary baseline because it represents a similarly lightweight adaptation approach - both methods introduce a small number of trainable parameters while keeping the base LLM frozen. 
Like our projector, soft prompts modify how the LLM processes inputs without changing its internal weights. 
This makes it a fair comparison point for assessing whether Vector-ICL's benefits come from the continuous vector representations themselves rather than just the additional training. 
For non-textual tasks, where soft prompts cannot be directly applied, we compare against cross-modal few-shot ICL and tuned encoders. 
In the non-textual domain, ICL inputs are represented either as numeric sequences (for time-series and brain fMRI data) or as textual descriptions (for graph edge lists and node features).
% }

\input{tex/tables/main_task_table}

\paragraph{Text Pretraining} 
To pretrain our text projectors, we leverage the WikiText-103~\citep{merity2016wikitext} dataset, consisting of over 100 million tokens from verified high-quality Wikipedia articles. 
This smaller language modeling corpus is chosen for its suitability to the lightweight nature of our projectors. The pretraining process is illustrated in~\cref{fig:pipeline_train}(a), where text snippets are divided at random sentence-end points. 
The first half is embedded and projected, while a next-token generation loss is applied to the second half.

\paragraph{Text Reconstruction} 
We investigate LLMs’ ability to decode original text from projected embeddings using two datasets: Parallel Sentence Talks~\citep{tiedemann2012parallelsetence} and Quora~\citep{thakur2021beir}. 
These datasets consist of concise text pieces that convey clear meaning. 
Projectors are trained on the training sets of both datasets, and performance is evaluated using the BLEU score~\citep{papineni2002bleu, sacrebleu}.

\paragraph{Arithmetic and Function Regression} 
For the arithmetic tasks, we generated synthetic datasets containing 10-digit numbers, which are particularly challenging for LLMs as they require splitting the numbers into multiple text tokens. These numbers are represented using a concatenated one-hot encoding per digit.
For instance, a 10-digit number is represented as a $10 \times 10$ matrix, flattened into a 100-dimensional vector.
The pretraining phase includes two key tasks: number reconstruction, where the model is tasked with recovering the original number from its embedding, and basic arithmetic, where the model performs algebraic addition operations on the projected embeddings. 

To evaluate the models' arithmetic reasoning abilities, we employ a non-linear function regression task, where the function is defined as \( f(x, y) = \sqrt{x} \sqrt{y} \). The model is provided with inputs \( x \) and \( y \), and it must predict the integer part of the function output. Performance is measured using the mean relative error, calculated as the \(\ell_1\) difference between the predicted and true values, normalized by the ground truth. This task allows us to assess the models' ability to perform more complex numerical reasoning beyond simple arithmetic operations.

\paragraph{Text Classification} 
We assess whether \methodlong can be applied effectively to text classification. 
Both binary and multi-class classification datasets are used, 
% as detailed in~\cref{tab:setup}, 
and the results are compared across few-shot ICL and soft prompt tuning.
% (\cref{tab:sentiment_analysis}). 
The classification performance is measured by accuracy.
% on the test set. 

\paragraph{Text Summarization} 
Following the classification tasks, we explore \methodlong's capability in summarizing text based on the projected embeddings. 
The datasets, encoders, LLMs, and prompt templates can be found in~\cref{tab:setup}.
Performance is evaluated using RougeL~\citep{lin-2004-rouge}. %against the ground-truth summaries.

\paragraph{Molecule Captioning}
We also extend our approach to the unconventional task of molecule captioning, using molecule sequence-caption pairs from the Language + Molecules-24 (LPM24)~\citep{edwards2024_LPM24} dataset. 
A sample molecule-caption pair looks like the following: 

\makebox[\textwidth][c]{
    \parbox{.9\textwidth}{
    Molecule: Cc1c(Cl)cccc1-n1ccn2c(SCC(=O)c3ccccc3C(F)(F)F)nnc2c1=O\\
    Caption: The molecule is a pain treatment that impacts inflammatory disease treatment.
    }
}

This task explores whether LLMs can extract useful information from projected embeddings of out-of-distribution chemical sequences, with performance evaluated via BLEU score.

\paragraph{Brain fMRI Decoding and Classification}
We analyze data from \citeauthor{lebel2022natural}~\citeyear{lebel2022natural} and \citeauthor{tang2023semantic}~\citeyear{tang2023semantic}, which consists of fMRI responses for 3 human subjects as they listen to 20+ hours of narrative stories from podcasts.
We preprocessed the data following \citeauthor{benara2024crafting}~\citeyear{benara2024crafting}, by converting the fMRI responses into a 200-dimensional output using principal components analysis and assigning classification labels to 10-grams of the story text at 2-second intervals using an LLM.
% GPT-4. 
% ensemble of LLMs.

% The data was separated into train set and test set by holding out the same three podcast stories from the three human subjects.
We use the same pretraining methodology as text to pretrain on the brain fMRI data (projecting on 20\% of time points and imposing next-token generation loss on the remaining 80\%) . 
% As the data comes in as segments of text and the recorded fMRI,
% We randomly sample 20\% of the fMRI timepoints in the training set and projected into box tokens, and we impose next token generation loss on the remaining 80\%. 
We evaluate the LLM's capability to decode projected brain fMRI by giving them randomly sampled context from the train set, that could come from different human subjects or from a different story, and ask them to decode segments from the test set. 
% \chandan{@Yufan is this correct? For this to work, the context should come from a different story but not from a different human subject.}
% \yufan{I see, running new experiments now with the correct configurations.}
% For fMRI decoding, we use this question, ``What is the English translation of the input?'' in the prompt template as shown in~\cref{tab:setup}. 

% We construct the classification questions around the properties of the underlying text, for example, ``Does the sentence contain a proper noun?'', ``"Does the input mention anything related to arguing?''.
% The ground truth is obtained via GPT4o~\citep{openai2023gpt4} as binary labels.

In addition to text reconstruction, we decode the binary labels from the fMRI responses, corresponding to questions about the underlying text, e.g. ``Does the sentence contain a proper noun?''
The decoding random baseline is constructed by giving the LLM the randomly sampled, shuffled text from the training set, and generating text according to it.
We measure the performance using the RougeL score between the generated text and the ground truth text.
The classification random baseline is 50\% accuracy, as we have balanced the dataset. 

\paragraph{Time-series}
We take the output of the last time step from Chronos-base~\citep{ansari2024chronos} as the time-series' representation.
We use the base encoder with trained classification head as the baseline and we measure the prediction performance with accuracy.

\paragraph{Graphs}
We use Graphormer~\citep{ying2021graphormer} as the encoder model, specifically the one that was pretrained on quantum chemistry graph datasets~\citep{hu2021ogb}.
Since the down-stream, ogbg-molhiv~\citep{hu-etal-2020-ogbg}, is a molecule property prediction dataset, and with strong class imbalance (3-4\% positive classes), we finetune the encoder on the training set to provide meaningful baselines and embeddings.
We take the output prior to the classification layer of the Graphormer as the graph embedding.
Weighted sampling is adopted in the finetuning of both the baseline Graphormer and the embedding projector to yield meaningful predictions. 
% We use the following prompt in ICL:
% \makebox[\textwidth]{$(\Box)$'s class (positive, negative) is: \underline{[Input Class]}}
We use the finetuned Graphormer as our baseline and use the F1 score as the performance metric.

\paragraph{Projector Configurations}
 Both linear and non-linear projectors are utilized, as shown in~\cref{fig:main_result}, with input and output dimensions matching the encoder-decoder pairs. 
 Early stopping with patience of 500 steps is used during finetuning, as projectors converge quickly due to their small parameter sizes. Details of the hyperparameters used in training are provided in~\cref{tab:hyperparameters}.

%% file: tex/tables/main_task_table.tex
% Define custom emoji commands for LLMs
\NewDocumentCommand\emojillama{}{\scalerel*{\includegraphics{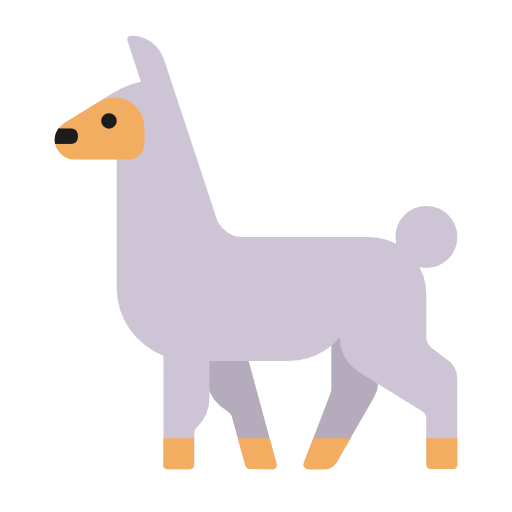}}{X}\xspace}
\NewDocumentCommand\emojimistral{}{\scalerel*{\includegraphics{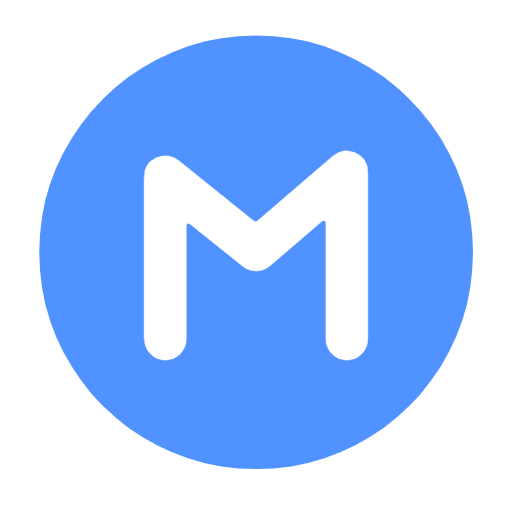}}{X}\xspace}
\NewDocumentCommand\emojiyi{}{\scalerel*{\includegraphics{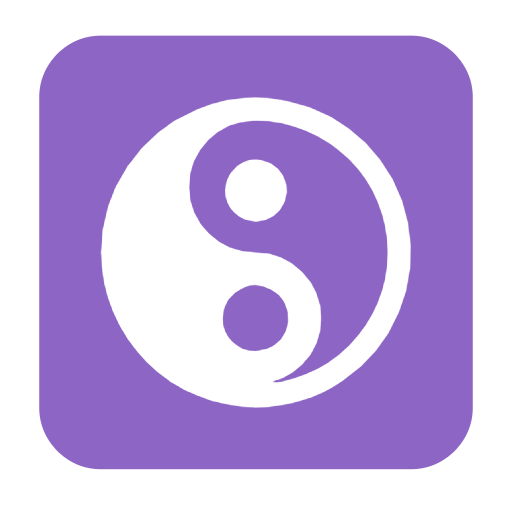}}{X}\xspace}
\NewDocumentCommand\emojiqwen{}{\scalerel*{\includegraphics{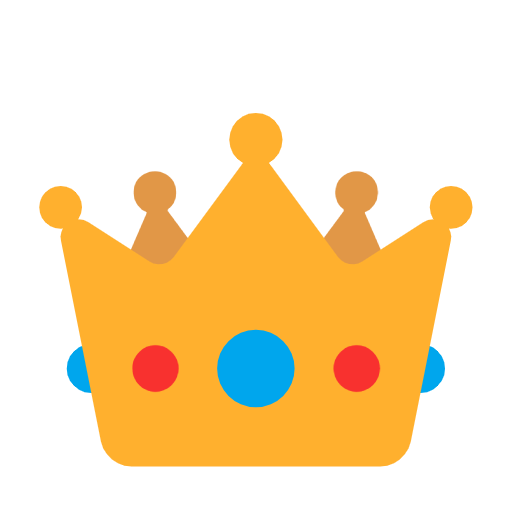}}{X}\xspace}

% Define custom emoji commands for Text Encoders
\NewDocumentCommand\emojinve{}{\scalerel*{\includegraphics{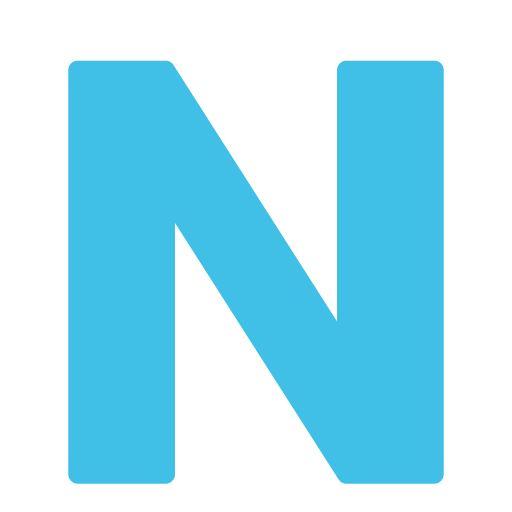}}{X}\xspace}
\NewDocumentCommand\emojisfr{}{\scalerel*{\includegraphics{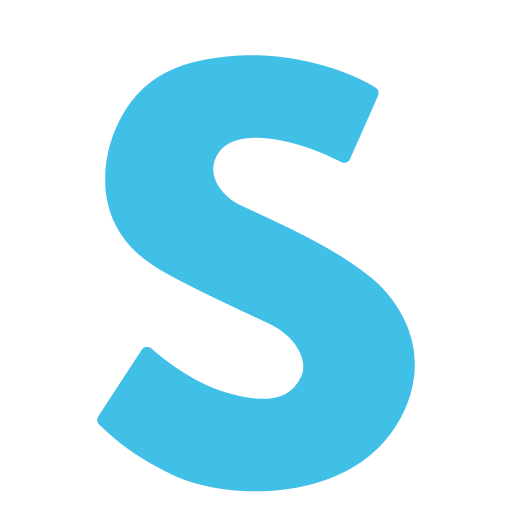}}{X}\xspace}
\NewDocumentCommand\emojistella{}{\scalerel*{\includegraphics{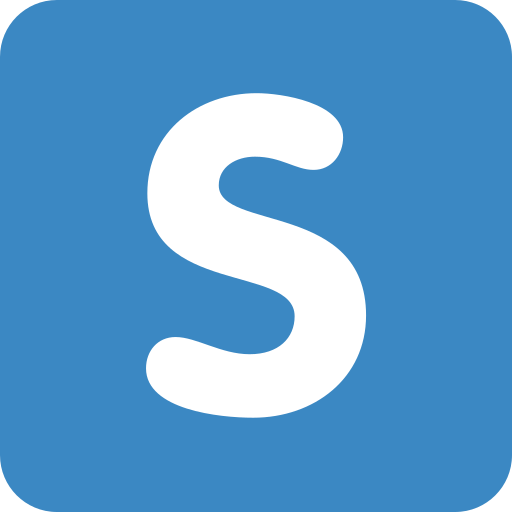}}{X}\xspace}
\NewDocumentCommand\emojigtr{}{\scalerel*{\includegraphics{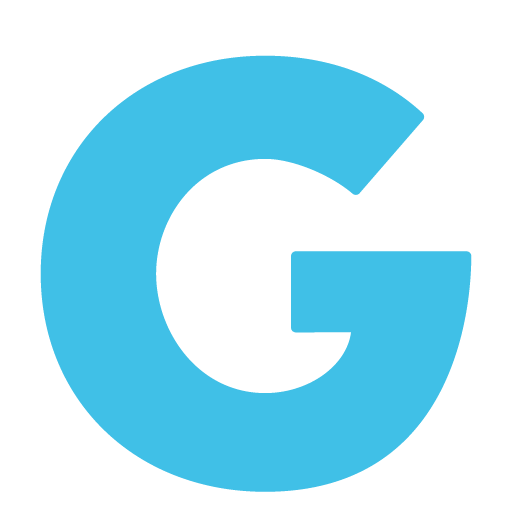}}{X}\xspace}

\renewcommand{\arraystretch}{1.5} % Increases the row height by 1.5 times
\begin{table}[t]
    
    \centering
    \caption{\textbf{Overview of the tasks, datasets, encoders, large language models, and prompts used in the experiments.} Each task utilizes encoders and LLMs to perform functions across multiple modalities. The table highlights the diversity of models and configurations applied to each task.}
    
    \scriptsize
    % \renewcommand{\arraystretch}{1.1}
    % \resizebox{\textwidth}{!}{
    \begin{tabularx}{\textwidth}{lllll}
    \toprule
    \textbf{Task} & \textbf{Dataset} & \textbf{Encoder} & \textbf{LLM} & \textbf{Prompt} \\ 
    \midrule
    
    Text Reconstruction 
        & \makecell[l]{Quora~\citep{thakur2021beir} \\ PST~\citep{tiedemann2012parallelsetence}}
        & \emojinve, \emojisfr, \emojistella, \emojigtr
        & \emojillama
        & \makecell[l]{Translate the text in brackets: $(\Box)$\\ Translation: [Text]} \\ \hline
    
    Function Regression  
        & 10 Digits Regression 
        & Digit Embedding
        & \makecell[l]{\emojillama, \emojimistral,\\ \emojiqwen, \emojiyi}
        & \makecell[l]{$x=\Box_x,\; y=\Box_y$,\\ function($x, y$) equals to (digits): [Solution]} \\ \hline
    
    Text Classification  
        & \makecell[l]{IMDB~\citep{maas-EtAl:2011:IMDB} \\ Rotten Tomatoes\\ \citep{Pang+Lee:05a:rotten_tomatoes}\\ SST2~\citep{socher-etal-2013-sst2} \\ Emotion~\citep{saravia-etal-2018-emotion}\\ Financial Phrasebank \\\citep{Malo2014financial_phrasebank}}
        & \emojinve, \emojisfr, \emojistella, \emojigtr
        & \makecell[l]{\emojillama,\\ \emojimistral,\\ \emojiqwen} %\emojiyi}
        & $(\Box)$'s sentiment is: [Label] \\ \hline
    
    Text Summarization   
        & \makecell[l]{XSum~\citep{Narayan2018xsum} \\ XLSum~\citep{hasan-etal-2021-xlsum}}
        & \emojinve, \emojistella, \emojigtr
        & \emojillama
        & $(\Box)$'s summarization is: [Summary] \\ \hline
    
    Molecule Captioning                
        & \makecell[l]{Language + Molecule-24\\ \citep{edwards2024_LPM24}}
        & \emojinve
        & \emojillama
        & \makecell[l]{$(\Box)$'s molecule caption is: [Caption]} \\ \hline
    
    Brain fMRI 
        & \makecell[l]{\citeauthor{lebel2022natural}~\citeyear{lebel2022natural}, \\ \citeauthor{tang2023semantic}~\citeyear{tang2023semantic}}
        & PCA of fMRI
        & \emojillama
        & \makecell[l]{[Question] Input: $\Box$ Response: [Answer]} \\ \hline
    
    \makecell[l]{Time-series\\ Classification}
        & \makecell[l]{FordA, FordB \\ \citep{dau2019ucrarchive}}
        & \makecell[l]{Chronos-base \\ \citep{ansari2024chronos}}
        & \emojillama
        & $(\Box)$'s class  (positive, negative) is: [Label] \\ \hline
    
    Graph Classification
        & \makecell[l]{ogbg-molhiv \\ \citep{hu-etal-2020-ogbg}}
        & \makecell[l]{Graphormer \\ \citep{ying2021graphormer}}
        & \emojillama, \emojiqwen
        & $(\Box)$'s class  (positive, negative) is: [Label] \\ \bottomrule
    \end{tabularx}
    % }
    \label{tab:setup}
    \begin{flushleft}
    \scriptsize
    \textbf{Encoders:} \emojinve\ NV-Embed~(\citeauthor{lee2024nvembed}~\citeyear{lee2024nvembed}; \texttt{nvidia/NV-Embed-v1}), \emojisfr\ SFR~(\citeauthor{meng2024sfrembedding}~\citeyear{meng2024sfrembedding}; \texttt{Salesforce/SFR-Embedding-2\_R}), \emojistella\ Stella~(\citeauthor{stella-en-v5}~\citeyear{stella-en-v5}; \texttt{dunzhang/stella\_en\_1.5B\_v5}), \emojigtr\ GTR-t5~(\citeauthor{ni2021gtr}~\citeyear{ni2021gtr}; \texttt{sentence-transformers/gtr-t5-base}) \\
    \textbf{LLMs:} \emojillama\ Llama-3.1-8B~(\citeauthor{dubey2024llama3}~\citeyear{dubey2024llama3}; \texttt{meta-llama/Llama-3.1-8B-Instruct}), \emojimistral\ Mistral-7B~(\citeauthor{jiang2023mistral}~\citeyear{jiang2023mistral}; \texttt{mistralai/Mistral-7B-Instruct-v0.3}), \emojiqwen\ Qwen2-7B~(\citeauthor{yang2024qwen2}~\citeyear{yang2024qwen2}; \texttt{Qwen/Qwen2-7B-Instruct}), \emojiyi\ Yi-1.5-9B~(\citeauthor{young2024yi}~\citeyear{young2024yi}; \texttt{01-ai/Yi-1.5-9B-Chat})
    \end{flushleft}
    \vspace{-3em}
\end{table}

\renewcommand{\arraystretch}{1}

%% file: tex/tables/merged_table.tex
\begin{table}[t]
    \centering
    \caption{Comparison of \methodlong, few-shot ICL, and soft prompt tuning across various sentiment analysis and summarization datasets. Details can be found in~\cref{sec:appendix_table_description}.}
    \label{tab:combined_results}
    \small
    \begin{tabular}{l |ccccc|cc}
    \toprule
    \textbf{Method} & \multicolumn{5}{c}{\textbf{Sentiment Analysis}} & \multicolumn{2}{c}{\textbf{Summarization}} \\
    \cmidrule(lr){2-6} \cmidrule(lr){7-8}
    & \makecell{\textbf{Rotten} \\ \textbf{Tomatoes}} & \textbf{SST2} & \textbf{IMDB} & \textbf{Emotion} & \makecell{\textbf{Financial} \\ \textbf{Phrasebank}} & \textbf{XSum} & \textbf{XLSum} \\
    \midrule
    \method (pretrained) & 80.60 & 78.90 & 95.04  & 41.20 & 60.72 & 15.25 & 15.89 \\
    Few-shot ICL      & 87.31 & 91.74 & 93.50 & 55.20 & 71.78 & 19.53 & 19.41 \\
    Soft Prompt  & \textbf{93.24} & 96.21 & 95.26  & 74.15 & 78.22 & 12.84 & 12.70 \\
    \method (finetuned)  & 88.80 & \textbf{98.16} & \textbf{97.28}  & \textbf{85.20} & \textbf{81.68} & \textbf{20.08} & \textbf{20.49} \\
    \bottomrule
    \end{tabular}
\end{table}

%% file: tex/s4_results.tex
\cref{fig:main_result} presents our main results, where each subplot corresponds to one of the nine tasks. 
We begin by exploring text reconstruction, to see whether LLMs can comprehend the information encoded within the box tokens. 
Next, we investigate the tasks of function regression to evaluate whether LLMs can leverage the box tokens during reasoning processes and whether this approach outperforms reasoning with plain text. 
Finally, we proceed to a range of downstream tasks, including text classification, text summarization, time-series classification, graph classification, and brain fMRI decoding.
This comprehensive evaluation allows us to assess the versatility and effectiveness of \methodlong across different domains and task types.

\paragraph{Text Reconstruction: LLM Understanding of Projected Embeddings}
We first verify LLMs' ability to understand projected embeddings. 
Our results demonstrate that \methodlong successfully reconstructs original text from projected embeddings, with performance improving as the number of examples (shots) increases, mirroring standard in-context learning (ICL) behavior. 
This suggests that LLMs can effectively decode the information compressed into the box tokens, with more context leading to better reconstruction. 
The similarity to ICL behavior indicates that \methodlong leverages similar learning mechanisms, but with the flexibility of working with continuous representations.

\paragraph{Function Regression: Enhanced Reasoning with Continuous Context}
We then study would it sometimes be better for LLMs to receive continuous context instead of discrete tokens.
After pretraining projectors on 10-digit number reconstruction and addition between two numbers, we task the LLM with learning an unknown function in-context. 
Results show \methodlong consistently outperforms few-shot ICL with raw number inputs, that have to span multiple tokens.
This suggests that the continuous representations capture numerical relationships more effectively than discrete tokens, enabling LLMs to better infer and apply mathematical patterns. 
The improvement is particularly noteworthy given that LLMs are typically challenged by precise numerical computations.

\paragraph{Text Classification}
In this classical NLP task, we aggregate mean accuracy across five datasets, four encoders, and three LLMs. 
Results indicate that pretrained projectors provide meaningful continuous context for ICL, outperforming the random baseline. 
LLMs achieve optimal performance with continuous context from finetuned projectors, surpassing both regular few-shot ICL and soft prompt tuning, with details shown in~\cref{tab:combined_results}. 
This demonstrates the versatility of \methodlong across different text classification scenarios and its ability to outperform established prompt tuning methods. 
The success across multiple datasets and LLMs suggests that the benefits of continuous context are robust and generalizable.

\paragraph{Text Summarization}
This task demands longer text generation and deeper information comprehension. We aggregate mean RougeL scores across two datasets, and three encoders. 
Findings show that pretrained projectors enable LLMs to extract and condense information from a single box token, while finetuned projectors provide more effective continuous context than original textual input and soft prompts, with details shown in~\cref{tab:combined_results}.
The ability to compress and later expand information from a single token is particularly impressive, suggesting that \methodlong captures high-level semantic content effectively. 
The superior performance of finetuned projectors highlights the benefits of task-specific optimization in continuous space.

\paragraph{Molecule Captioning: An Unconventional NLP Task}
We explore LLMs' ability to comprehend continuous vector context for out-of-distribution inputs like molecule sequences. 
Evaluating captioning performance with BLEU scores, we find that both pretrained and finetuned projectors provide better context than original molecule sequence text, despite encoders likely never trained on such sequences. 
This result is particularly intriguing as it demonstrates \methodlong's ability to bridge the gap between specialized domains and general language understanding. 
It suggests that continuous representations can capture and translate domain-specific information in a way that's more accessible to LLMs than raw specialized notation.

\paragraph{Time-series Classification}
We finetune non-linear projectors on the training sets of two datasets and evaluate LLM performance with the resulting continuous context. 
Aggregating average accuracy, we find LLMs outperform baseline domain-specific models with finetuned classification heads. 
% This result is significant as it shows that \methodlong can adapt LLMs to tasks typically handled by specialized architectures. 
The success here suggests that continuous context can effectively capture temporal dependencies and patterns, translating time-series data into a form that LLMs can process effectively.

\paragraph{Graph Classification}
Since the task dataset is heavily imbalanced and out of the pretraining distribution of the graph encoder.
We first finetune the base encoder on the target dataset to establish a baseline, then train non-linear projectors on the graph classification dataset. 
Averaging F1 scores across two LLMs, results indicate that \methodlong enables LLMs to outperform the finetuned baseline model. This is a noteworthy achievement, as graph data is structurally very different from the text data that LLMs are trained on. 
The success here suggests that \methodlong can effectively translate graph structures into continuous representations that preserve relational information in a way that's interpretable to LLMs.

\paragraph{Brain fMRI Decoding and Classification}
We pretrain projectors on the training set using next-token generation loss, then apply them to recover original text from brain fMRI signals in the test set. 
Results show LLMs can surpass random baselines by leveraging projected 200-dimensional fMRI PCA factors, with performance improving as context increases. 
This application to neuroscience data is particularly exciting, demonstrating \methodlong's potential in bridging neural activity and language understanding. 
% The improvement with increased context suggests that LLMs can aggregate information across multiple fMRI readings, potentially capturing temporal or spatial patterns in brain activity.

%% file: tex/s4.5_analysis.tex
\subsection{The Impact of Encoder Quality on \methodlong Performance}
We investigate the relationship between the intrinsic capabilities of encoders and their effectiveness when used with \methodlong for downstream tasks. To quantify the encoders' intrinsic abilities, we evaluate their performance on a text reconstruction task, which serves as a proxy for the amount of information preserved in the embeddings.

Our analysis focuses on text classification as the downstream task. We examine the correlation between encoder rankings on the reconstruction task and their corresponding rankings on the classification task. This analysis is performed across 5 datasets and 3 LLMs, resulting in 15 configurations.

The results of this analysis, presented in \cref{fig:encoder_analysis}, demonstrate a consistent positive correlation between an encoder's text reconstruction performance and its effectiveness in downstream classification tasks when used with \methodlong. Notably, in 4 of the 15 configurations, we observe a particularly strong correlation, with values approaching 1.

Our findings suggest that an encoder's performance on the text reconstruction task can serve as a reliable predictor of its potential effectiveness in downstream tasks when integrated with \methodlong. This insight could prove valuable for practitioners in selecting encoders for~\methodlong.

% Future work could explore whether this relationship holds for other types of downstream tasks beyond text classification, and investigate the specific characteristics of encoders that contribute to both strong reconstruction performance and downstream task effectiveness.
% \input{tex/figures/analysis_encoder} 

\subsection{Case study: what has been learned in the projections?}

\cref{fig:number analysis} provides a visualization of the normalized Euclidean distances between projected embeddings. %where the x and y axes correspond to 1024 different numbers uniformly sampled from $0$ to $1e10$ and the color intensity reflects the distance between each pair. 
Several key patterns emerge from this that offer insights into what the projector has learned.

Analysis of the numerical embedding distance matrix reveals key properties of our projection method. Embeddings for similar numbers cluster along the diagonal, indicated by lighter colors, demonstrating the preservation of local structure. Conversely, increasing distances from the diagonal, shown by darker colors, indicate effective separation of numerically distant values in the embedding space.

Another notable feature of the distance matrix is the block structure that emerges. 
% \textcolor{blue}{
The block structure reflects how numbers share similar digit patterns across the decimal places, from local to global blocks.
It likely helps LLMs process numerical relationships more effectively, as it preserves the hierarchical nature of place-value notation.
%}

%This pattern comes from the way we construct the numerical embeddings and it is likely beneficial for the LLM to understand numerical inputs.

% Furthermore, the distance matrix reveals smooth transitions in distance values as we move from the diagonal outward. 
% Rather than sharp boundaries between projections, the projector seems to have learned a continuous mapping of numbers, implying that it captures both the magnitude and structure of the input in a smooth, interpretable way. 
% This continuity is important for enabling the model to perform reasoning tasks, as it ensures that similar inputs maintain similar embeddings.

\input{tex/figures/analysis_numbers}

\subsection{Case study: can synthetic dataset be useful in cross-modal pretraining?}

The cross-modal pretraining requires data-text dual pairs, which is not available in many cases. 
We explore how to create a synthetic dataset for such pretraining using time-series as an example. 
We designed a data curation pipeline that extracts meaningful statistical properties from time-series and converts them into natural language descriptions.

Our pipeline analyzes multiple statistical aspects of the time-series data. We perform trend analysis through linear regression to capture overall directional movements, anomaly detection using z-score analysis to identify unusual patterns, and temporal stability assessment via variability thresholds to characterize consistency.

For each identified property, we generate both descriptive statements and binary questions. For instance, given a time-series segment, we generate descriptive text like "The time-series shows an upward trend" or binary questions such as "Does this time-series contain any anomalies?"

Our experiments show that projectors pretrained on this synthetic corpus can effectively generalize to new tasks, demonstrating the feasibility of creating synthetic data for cross-modal pretraining when natural data pairs are unavailable.

%% file: tex/figures/analysis_numbers.tex
\begin{figure}[t]
    \centering
    \resizebox{\linewidth}{!}{
    \begin{subfigure}[t]{0.3\textwidth}
        \vspace{-1em}
        \caption{}
        \adjustbox{valign=t, keepaspectratio, max width=\textwidth}{\includegraphics{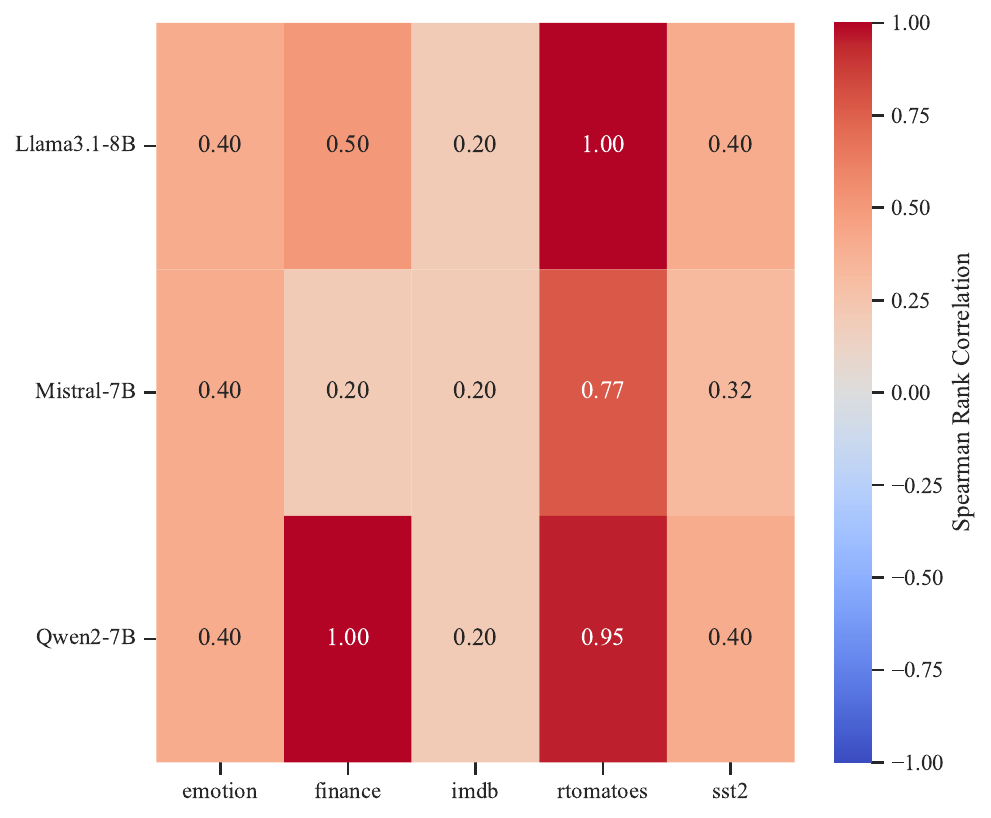}}
        
        \label{fig:encoder_analysis}
    \end{subfigure}
    \hspace{0.01\textwidth}
    \begin{subfigure}[t]{0.3\textwidth}
        \vspace{-1em}
        \caption{}
        \adjustbox{valign=t, keepaspectratio, max width=\textwidth}{\includegraphics{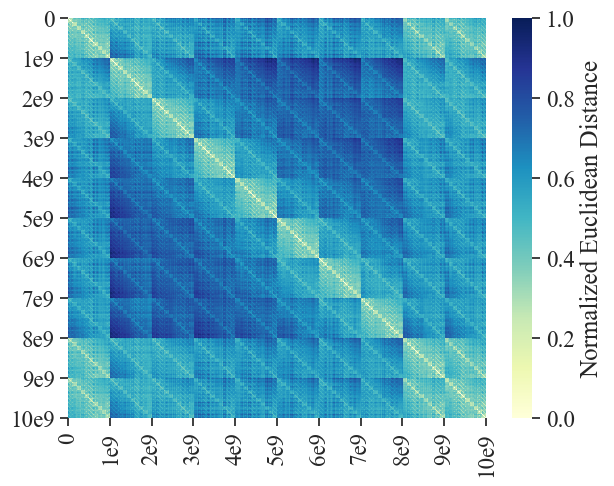}}
        
        \label{fig:number analysis}
    \end{subfigure}
    \hspace{0.01\textwidth}
    \begin{subfigure}[t]{0.3\textwidth}
        \vspace{-1em}
        \caption{}
        \adjustbox{valign=t, keepaspectratio, max width=\textwidth}{\includegraphics{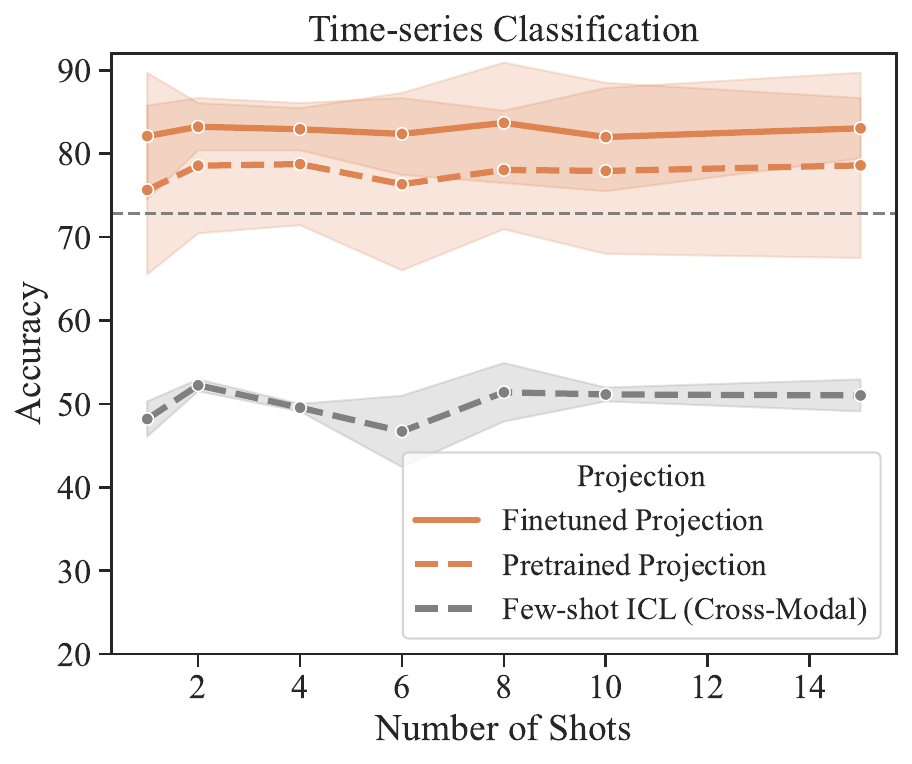}}
        
        \label{fig:time series pretrain}
    \end{subfigure}
    }
    \caption{\textbf{Key Insights from Encoders, Projections, and Synthetic Data Curation.} 
    (a) Correlation between encoders' text reconstruction performance and their downstream task effectiveness with \methodlong, suggesting information preservation ability predicts Vector-ICL performance.
    (b) Euclidean distance matrix of 1024 projected number embeddings (0 to 1e10) shows structured block-diagonal patterns, indicating meaningful numerical relationships are preserved.
    % \textcolor{blue}{
    (c) Results from pretraining on our synthetic time-series QA dataset, which captures statistical properties through trend analysis, anomaly detection, and stability assessment. The curated QA pairs enable effective cross-modal pertaining.
    %}
    }
    \label{fig:projector_analysis}
\end{figure}

%% file: tex/s5_discussion.tex
\paragraph{Limitations and Future Directions}
In this study, we explored a variety of settings: utilizing different encoders, LLM architectures, modalities, and datasets. 
Our results demonstrate that LLMs are capable of performing \methodlong on both language and non-language inputs. 
However, our experiments did not cover all possible combinations of these variables. 
There are still many unexplored areas, such as additional modalities, tasks, and encoder-decoder configurations, that could further benefit from \methodlong.
Also, we only experimented with single-token encoders, while there exist encoders that produce variable-sized embeddings, that can potentially be more powerful and flexible.
% \textcolor{blue}{
Analyzing how instruction tuning might affect the model's ability to understand vector context would be beneficial as well.
% }
We leave this extensive exploration for future research to fully understand the broader applicability and limitations of our approach across diverse domains.

\paragraph{Conclusion}
In this work, we explore whether large language models trained only on text can perform in-context learning on continuous vectors from different domains.
Our findings suggest that LLMs can indeed understand and process continuous context via embedding projection. 
Simple linear projections are often sufficient, though for cross-modal tasks—such as those involving non-textual data like time-series or graphs—non-linear transformations may be required.
% We demonstrate that training the embedding projector using a straightforward next-token prediction objective enables~\methodlong. 
In our experiments across various tasks and modalities, including text reconstruction, numerical function regression, text classification, summarization, molecule captioning, time-series classification, graph classification, and fMRI decoding, Vector-ICL often surpasses both few-shot ICL and task-specific model or tuning. 
We further conduct analyses and case studies, indicating the potential of LLMs to process vector representations beyond traditional token-based paradigms.

%% file: tex/appendix.tex
\section{Appendix}
\subsection{Detailed Experiment Setup}
\input{tex/s3.5_setup_appendix}

\input{tex/figures/brain_category_plot}
\subsection{Case study: what information was perceived from projected brain fmri?}
\label{sec:output_analysis_brain}

\cref{fig:brain question} illustrates the mean accuracy achieved in decoding different categories of brain activity based on fMRI data. 
The categories, ranging from "Physical Actions and Movements" to "Conflict, Urgency, and Change," represent diverse cognitive and perceptual domains.
The grouping and corresponding questions are listed in~\cref{tab:question_categories}.

The input data is noisy, and the projector is only trained with pretraining objectives, i.e., to predict the next piece of text given the current fMRI signal.
We are surprised that with this pure unsupervised training, the LLM can still pick up meaningful signals from the projected embeddings.
Notably, decoding tasks associated with ``Physical Actions and Movements'' and ``Cognitive and Reflective Aspects'' demonstrate higher mean accuracy, suggesting that these categories are more distinguishable based on the fMRI embeddings. 

\subsection{Text classification and text summarization config in tables}
\label{sec:appendix_table_description}
Llama3.1-8B is used as the common LLM and NV-Embed-v1 is used as the text encoder for \methodlong. The ICL and soft-prompt methods are supplied with up to 50 shots as context. The soft-prompt tokens are trained over the entire training set.

\clearpage
\subsection{Additional Experimental Results}
\input{tex/figures/reconstruction_figure}
% \subsection{Function regression results}
\input{tex/figures/digits}

% \subsection{Text classification results}
\input{tex/figures/text_classification_pretrained}
\input{tex/figures/text_classification_finetuned}
\input{tex/figures/text_classification_fewshot}

\input{tex/figures/text_summarization_pretrained}
\input{tex/figures/text_summarization_finetuned}

% \subsection{Hyperparameters}
\input{tex/tables/training_config}

\clearpage
\subsection{Brain fmri question categories}
\input{tex/tables/brain_question_table}

%% file: tex/s3.5_setup_appendix.tex
\label{sec:appendix_setup}
\paragraph{Text Reconstruction}
We use two datasets for the text reconstruction task, Parallel Sentence Talks's English subset~\citep{tiedemann2012parallelsetence} and Quora~\citep{thakur2021beir}.

Parallel Sentence Talks consist of sentences used in movie conversations, and Quora is built on a wide range of online questions.
They represent short pieces of text that convey clear information. 
We aim to explore whether LLMs can decode the original message from the projected text embeddings.

We use the following prompt template:

\makebox[\textwidth]{Translate the text in brackets: $(\Box)$, translation: \underline{[Original Text]}}

We train the projectors on the training set of these two datasets and evaluate their performance on the test tests.
We measure the reconstruction performance with the BLEU score~\citep{papineni2002bleu, sacrebleu}.

\paragraph{Arithmetic and Function Regression}
We created synthetic datasets of numerical data to pretrain our linear number projectors, experimenting with two configurations: one using 3-digit numbers and the other using 10-digit numbers. In Llama3.1-8B’s tokenizer, 3-digit numbers are represented as single tokens, while in Mistral-7B, Qwen2-7B, and Yi-1.5-9B, numbers larger than 10 are split into multiple tokens. Consequently, 10-digit numbers consistently span multiple tokens across all models, which increases the complexity of performing arithmetic operations.

To represent the numbers, we use a concatenated-and-flattened one-hot vector encoding for each digit. 
For instance, a 3-digit number is represented as a $3 \times 10$ matrix (one hot per digit place), which is then flattened into a 30-dimensional vector. 
Similarly, a 10-digit number is represented as a $10 \times 10$ matrix, flattened into a 100-dimensional vector.

The pretraining involves two tasks. 
The first task is number reconstruction, we use the following prompt template, given the number is 128:

\makebox[\textwidth]{$x=\Box_x$,\quad $x$  equals to (digits): \underline{128}}

The second task is basic addition, we use the following prompt template, given the numbers are $x=128\;y=256,\;a=1,\;b=1$:

\makebox[\textwidth]{$x=\Box_x,\; y=\Box_y$,\quad $a*x+b*y$ equals to (digits): \underline{384}}

Here, $a$ and $b$ are randomly sampled from $[0,1]$ with up to two decimal places, and the solution is the integer part of the sum.

For evaluation, we use a function regression task with a non-linear function: $f(x, y) = \sqrt{x} \cdot \sqrt{y}$. 
The LLM is given inputs $x$ and $y$, along with the integer part of the output $f(x, y)$. 
The model is then tasked with predicting the output for new pairs of $x$ and $y$. 
The prompt for in-context learning is structured as follows:

\makebox[\textwidth]{$x=\Box_x,\; y=\Box_y$,\quad function($x, y$) equals to (digits): \underline{[Solution]}}

For few-shot ICL, the box tokens will be replaced with the actual numbers.
We measure the function regression performance with the mean relative error, where the relative error is computed as the $\ell_1$ difference divided by the ground truth value.

\paragraph{Text Classification}
We use five datasets for the text classification task. 
For binary classification, we include IMDB~\citep{maas-EtAl:2011:IMDB}, Rotten Tomatoes~\citep{Pang+Lee:05a:rotten_tomatoes}, and the Stanford Sentiment Treebank (SST2)~\citep{socher-etal-2013-sst2}. 
For multi-class classification, we use the Emotion dataset~\citep{saravia-etal-2018-emotion} and the Financial Phrasebank~\citep{Malo2014financial_phrasebank}.
The binary classification datasets (IMDB, Rotten Tomatoes, and SST2) involve classifying movie reviews as positive or negative. 
The Emotion dataset classifies Twitter tweets into six categories: anger, fear, joy, love, sadness, and surprise. 
The Financial Phrasebank categorizes financial news into positive, negative, or neutral sentiments.

We use the following prompt in ICL:

\makebox[\textwidth]{$(\Box)$'s sentiment is \underline{[Input Class]}}

For few-shot ICL, the box tokens will be replaced with the actual text. For soft prompt tuning, we use 20 virtual tokens and train for one epoch over the entire training set.
We measure the classification performance with accuracy on the test set.

\paragraph{Text Summarization}
We use two datasets for the summarization task, XSum~\citep{Narayan2018xsum} and the English subset of XLSum~\citep{hasan-etal-2021-xlsum}.
They contain newspaper articles and their summaries.

We use the following prompt in ICL:

\makebox[\textwidth]{$(\Box)$'s summarization is: \underline{[Summary of the Input]}}

For few-shot ICL, the box token will be replaced by the article. 
We measure the performance using the RougeL score with the ground truth summary on the test sets.

\paragraph{Molecule Captioning}
We use the Language + Molecules-24 (LPM24) dataset for the molecule captioning task, it was created for the task of molecule-language translation, and contains 161K pairs of molecule strings and their captions in the combined training and test set.

A sample molecule-caption pair looks like the following: 

\makebox[\textwidth][c]{
    \parbox{.9\textwidth}{
    Molecule: Cc1c(Cl)cccc1-n1ccn2c(SCC(=O)c3ccccc3C(F)(F)F)nnc2c1=O\\
    Caption: The molecule is a pain treatment that impacts inflammatory disease treatment.
    }
}

And we use the following prompt for ICL:

\makebox[\textwidth]{$(\Box)$'s molecule caption is: \underline{[Caption of the Input Molecule]}}

For few-shot ICL, we replace the box token with the actual molecule string.
We measure the performance using the BLEU score between the generated caption and the ground truth caption.

\paragraph{Brain fMRI Decoding and Classification}
We analyze data from \citeauthor{lebel2022natural}~\citeyear{lebel2022natural} and \citeauthor{tang2023semantic}~\citeyear{tang2023semantic}, which consists of fMRI responses for 3 human subjects as they listen to 20+ hours of narrative stories from podcasts.
We preprocessed the data following \citeauthor{benara2024crafting}~\citeyear{benara2024crafting}, by converting the fMRI responses into a 200-dimensional output using principal components analysis and labeling 10-grams of the story text at 2-second intervals using an ensemble of LLMs.

The data was separated into train set and test set by holding out the same three podcast stories from the three human subjects.
We use the same pretraining methodology as text to pretrain on the brain fMRI data. 
As the data comes in as segments of text and the recorded fMRI, we randomly sample 20\% of the segments to be in fMRI form and projected into box tokens, and we impose next token generation loss on the rest 80\%. 

We evaluate the projectors by giving them randomly sampled context from the train set, that could come from different human subjects or from a different story, and ask them to decode segments from the test set.
We use the following prompt in ICL in our decoding experiments:

\makebox[\textwidth]{
    \parbox{0.75\textwidth}{
    What is the English translation of the input?\\ 
    Input: $\Box$, Response: the input in English is \underline{[Text Corresponding to fMRI]}}
}

The random baseline is constructed by giving LLM the randomly sampled, shuffled text from the training set, and generating text according to it.
We measure the performance using the RougeL score between the generated text and the ground truth.

We construct the classification questions around the properties of the underlying text, for example, ``Does the sentence contain a proper noun?'', ``"Does the input mention anything related to arguing?''.
The ground truth is obtained via GPT4o~\citep{openai2023gpt4} as binary labels.
We use the following prompt in ICL in our classification experiments, using one of the example questions:

\makebox[\textwidth]{
    \parbox{0.75\textwidth}{
    Does the input mention anything related to arguing?\\ 
    Input: $\Box$, Response (Yes or No): according to the English text of the input, the answer is \underline{[Yes/No]}}
}

The random baseline is 50\%, as we have downsampled and balanced the data. And we use accuracy as the performance metric.

\paragraph{Time-series}
We use the Chronos~\citep{ansari2024chronos} time-series Transformers as the encoder. 
Chronos was pretrained on large scale time-series and is designed to generate the next segments of the time-series.
It has proven effective on a wide range of time-series forecasting benchmarks. 
We take the output of the last time step from Chronos-base as the time-series representation.

We use two datasets for the time-series classification task, FordA, and FordB, they are also part of the UCR Time Series Classification Archive~\citep{dau2019ucrarchive} ranging from 4000 to 5000 time-series for each dataset.
We use the following prompt in ICL:

\makebox[\textwidth]{$(\Box)$'s class (positive, negative) is: \underline{[Input Class]}}

We construct a synthetic pretraining corpus through comprehensive analysis of additional time-series datasets from the UCR Time Series Classification Archive \citep{dau2019ucrarchive}, specifically MoteStrain, TwoLeadECG, Wafer, PhalangesOutlinesCorrect, and Yoga. Our data curation framework includes multiple statistical approaches: trend identification via linear regression, anomaly detection using z-score analysis, and assessment of temporal stability through variability thresholds. To enhance corpus diversity, we incorporate binary (yes/no) questions targeting specific time-series characteristics. The resulting QA pairs create a rich textual representation that captures the underlying temporal and statistical properties of the data.

We use the base encoder with trained classification head as the baseline and we measure the prediction performance with accuracy.

\paragraph{Graphs}
We use Graphormer~\citep{ying2021graphormer} as the encoder model, specifically the one that was pretrained on large scale quantum chemistry graph datasets~\citep{hu2021ogb}.
Since the down-stream task (ogbg-molhiv~\citep{hu-etal-2020-ogbg}) is a molecule property prediction dataset, and with strong class imbalance (3\% positive classes), we finetune the encoder on the training set to provide meaningful baselines and embeddings.
We take the output prior to the classification layer of the Graphormer as the graph embedding.

We use the ogbg-molhiv~\citep{hu-etal-2020-ogbg} dataset for the graph classification task. 
ogbg-molhiv is a molecule property prediction dataset consisting of a total 41.12K graphs with node features and edge attributes.
It has a strong class imbalance of having around 3\% positive class and 97\% negative class.

Hence weighted sampling is adopted in the finetuning of both the baseline Graphormer and the embedding projector to yield meaningful predictions. 
We use the following prompt in ICL:

\makebox[\textwidth]{$(\Box)$'s class (positive, negative) is: \underline{[Input Class]}}

We use the finetuned Graphormer as our baseline and use the F1 score as the performance metric due to the significant class imbalance.

% % \input{tex/tables/projector_task_table}
% \paragraph{Projectors Configurations}
% We have linear and non-linear projectors, and we used the configuration for each modality as described in~\cref{fig:main_result}.
% The input and output dimensions of the projectors are configured to match the output dimension of the encoder and the input dimension of the decoder, respectively.
% We train the projectors one epoch over the training set of the specific task in both pretraining and finetuning. 
% In finetuning, we further set up early stopping with max patience equal to 500 steps, since the projectors quickly become convergent because of their small parameter size.
% The details of training are listed in~\cref{tab:hyperparameters}.

%% file: tex/figures/brain_category_plot.tex
\begin{figure}
    \begin{center}
    \includegraphics[width=0.4\linewidth, keepaspectratio]{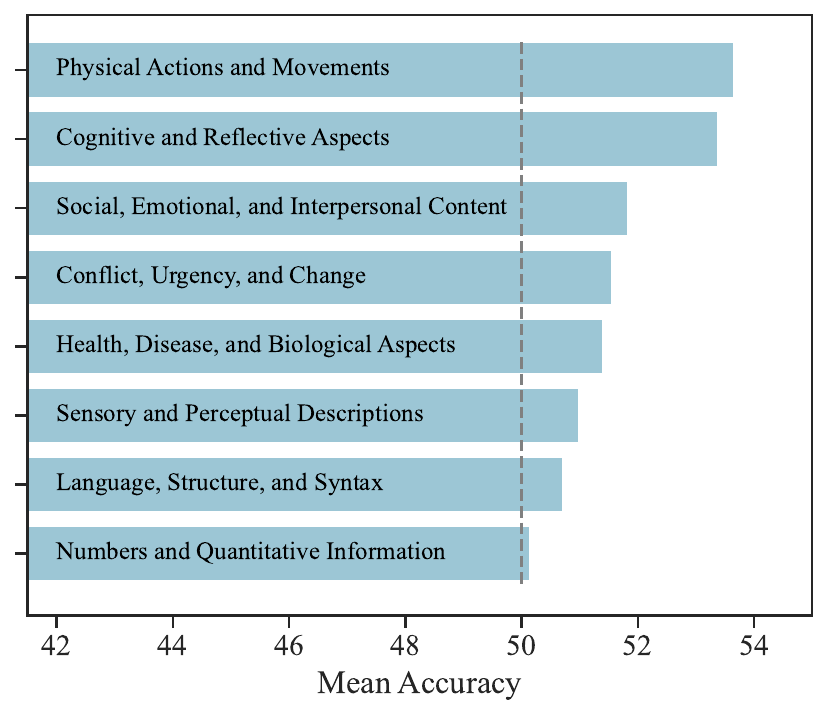}
    \end{center}
    \caption{Analyzing LLM's understanding of projected brain fMRI embeddings after only unsupervised pretraining. We categorize the underlying questions related to the text and measure the mean accuracy for each category, highlighting the LLM’s ability to interpret the embeddings, with only the next token prediction pretraining.}
    \label{fig:brain question}
\end{figure}

%% file: tex/figures/reconstruction_figure.tex
\begin{figure}[ht]
    \centering
    \begin{subfigure}[t]{0.45\linewidth}
        \centering
        \includegraphics[height=5cm, keepaspectratio]{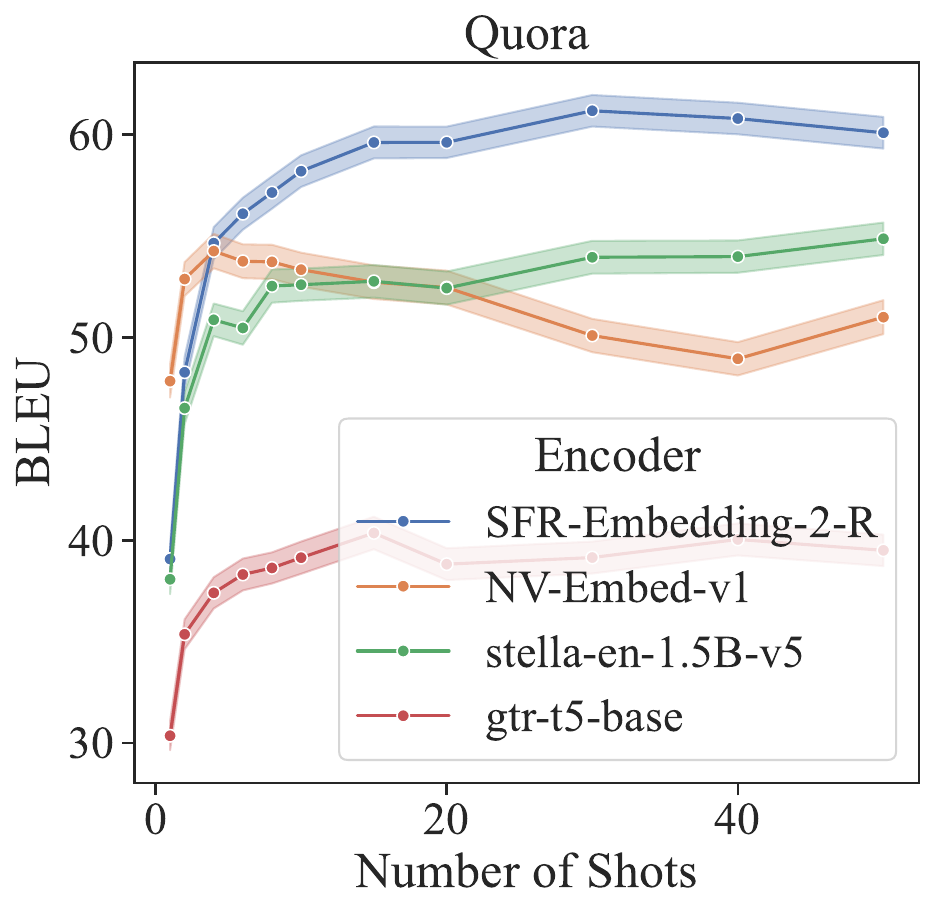}
        \label{fig:reconstruction_quora}
    \end{subfigure}\hfill
    \begin{subfigure}[t]{0.45\linewidth}
        \centering
        \includegraphics[height=5cm, keepaspectratio]{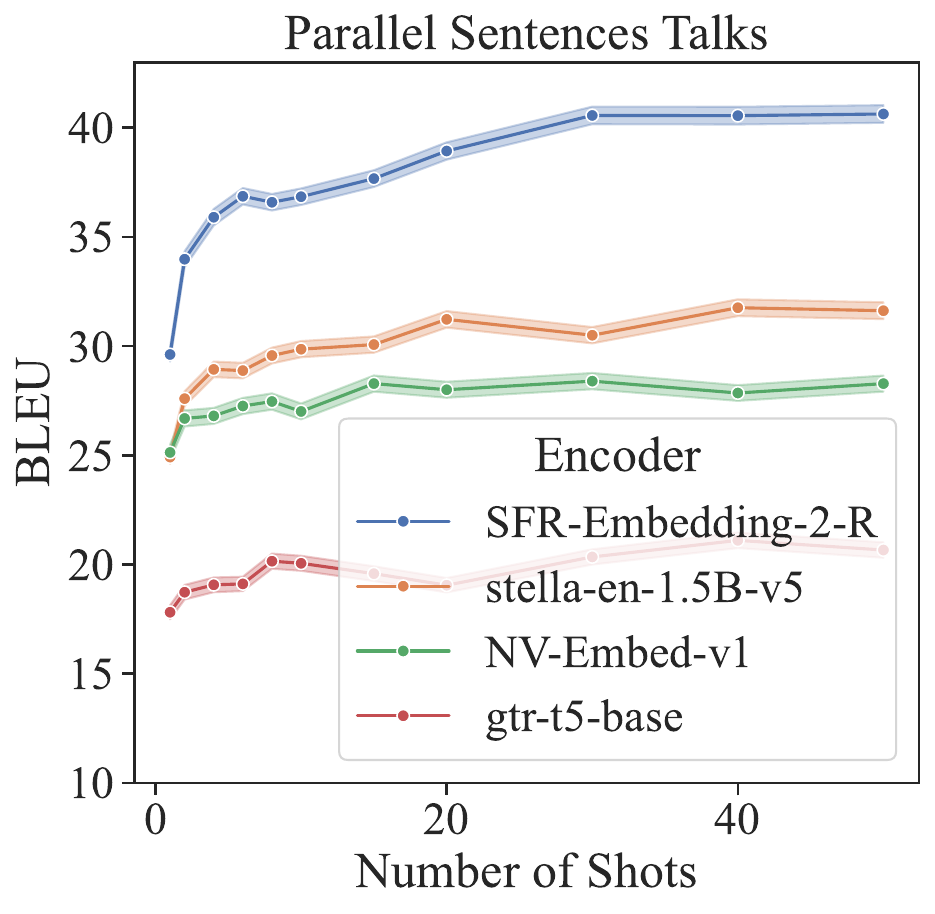}
        \label{fig:reconstruction_pst}
    \end{subfigure}
    \caption{Text Reconstruction}
    \label{fig:text reconstructino}
\end{figure}

%% file: tex/figures/digits.tex
\begin{figure}
    \centering
    \begin{subfigure}[b]{\textwidth}
        \centering
        \includegraphics[keepaspectratio, width=\textwidth]{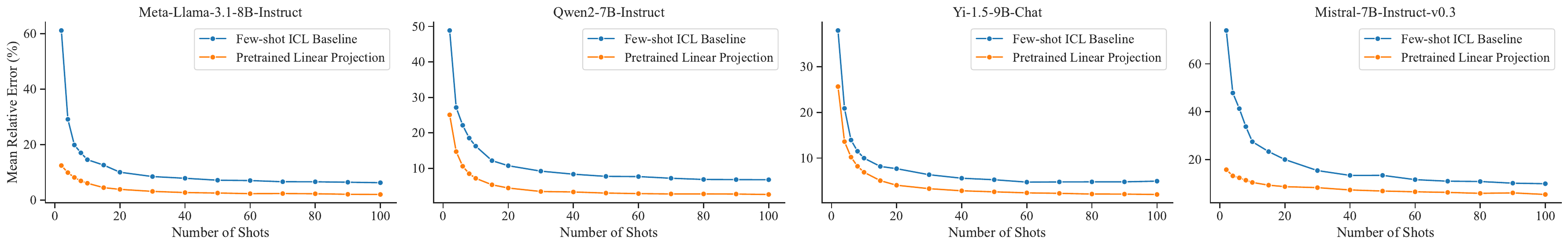}
    \end{subfigure}
    \begin{subfigure}[b]{\textwidth}
        \centering
        \includegraphics[keepaspectratio, width=\textwidth]{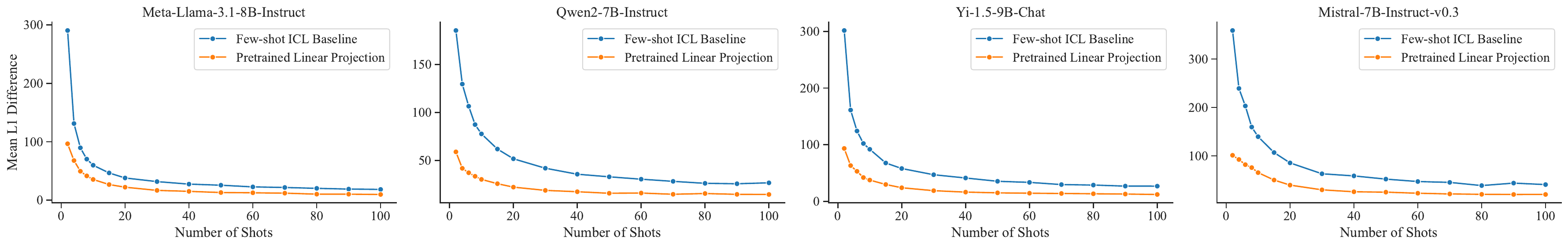}
    \end{subfigure}
    \caption{Function Regression - 10 digits (upper) and 3 Digits (lower)}
\end{figure}

%% file: tex/figures/text_classification_pretrained.tex
\begin{figure}[ht]
    \centering
    \includegraphics[keepaspectratio, width=\textwidth]{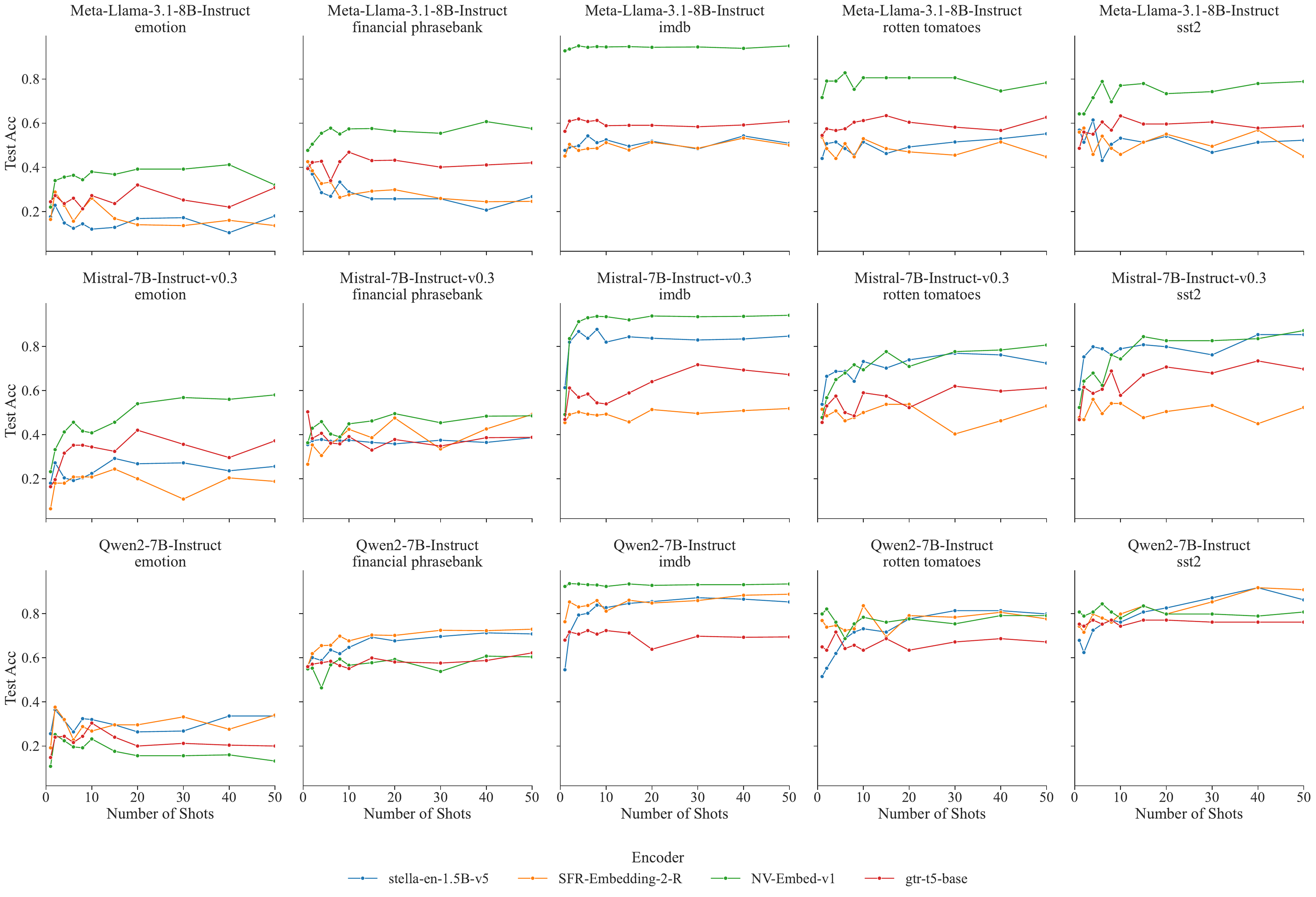}
    \caption{Text Classification - Pretrained Projectors}
\end{figure}

%% file: tex/figures/text_classification_finetuned.tex
\begin{figure}[ht]
    \centering
    \includegraphics[keepaspectratio, width=\textwidth]{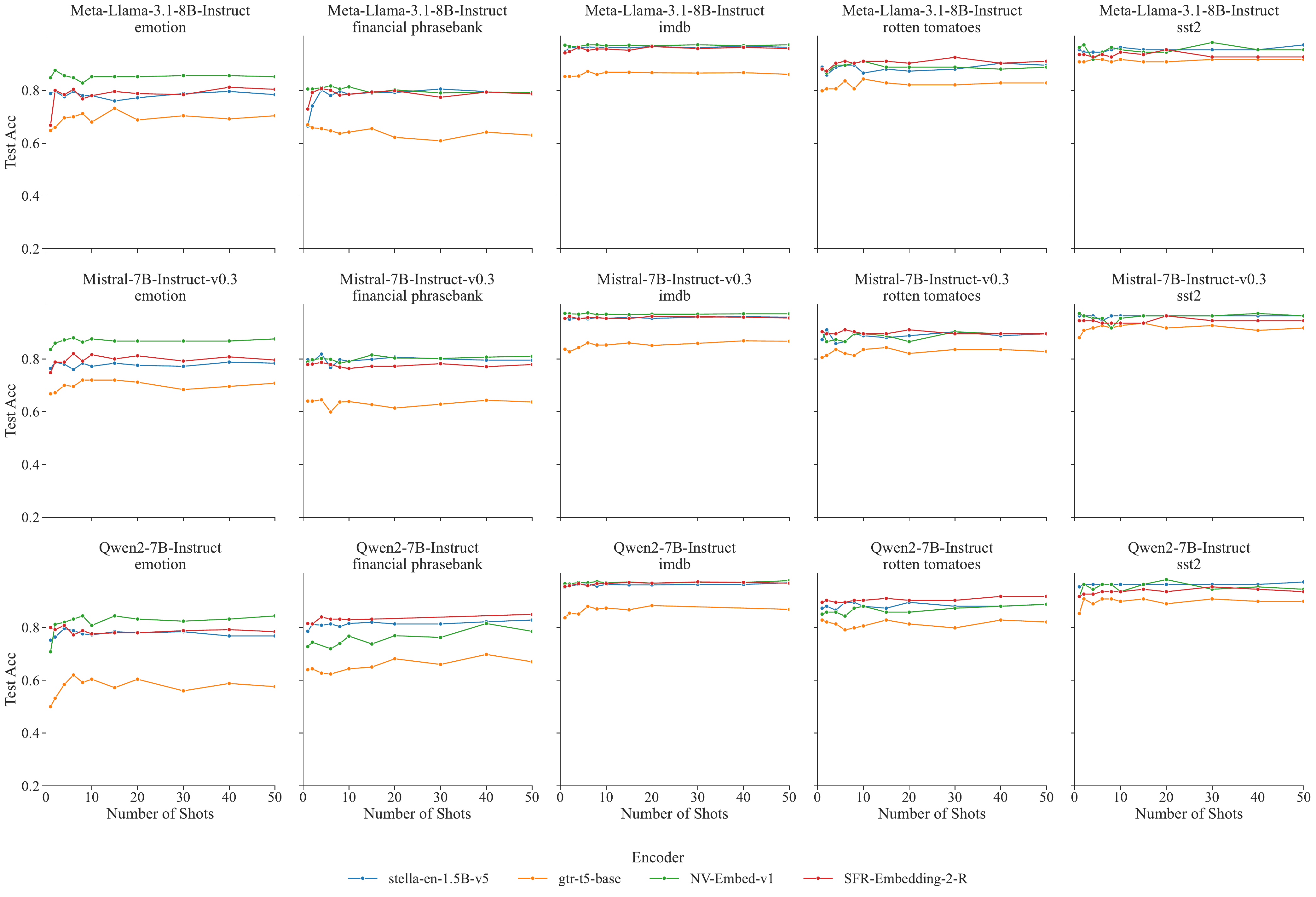}
    \caption{Text Classification - Finetuned Projectors}
\end{figure}

%% file: tex/figures/text_classification_fewshot.tex
\begin{figure}[ht]
    \centering
    \includegraphics[keepaspectratio, width=\textwidth]{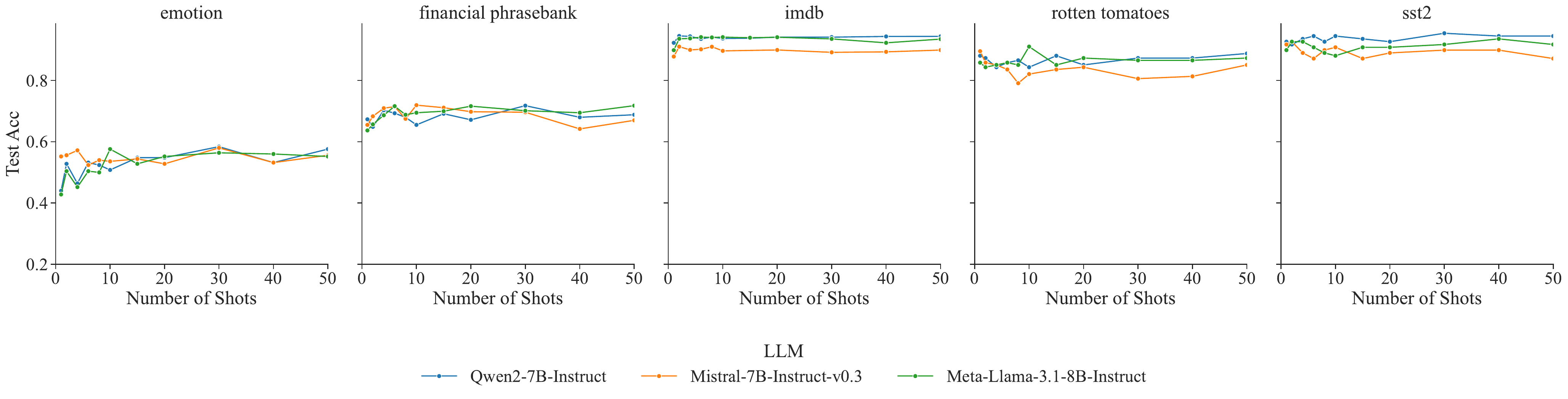}
    \caption{Text Classification - Few-shot ICL}
\end{figure}

%% file: tex/figures/text_summarization_pretrained.tex
\begin{figure}[ht]
    \centering
    \includegraphics[keepaspectratio, width=\textwidth]{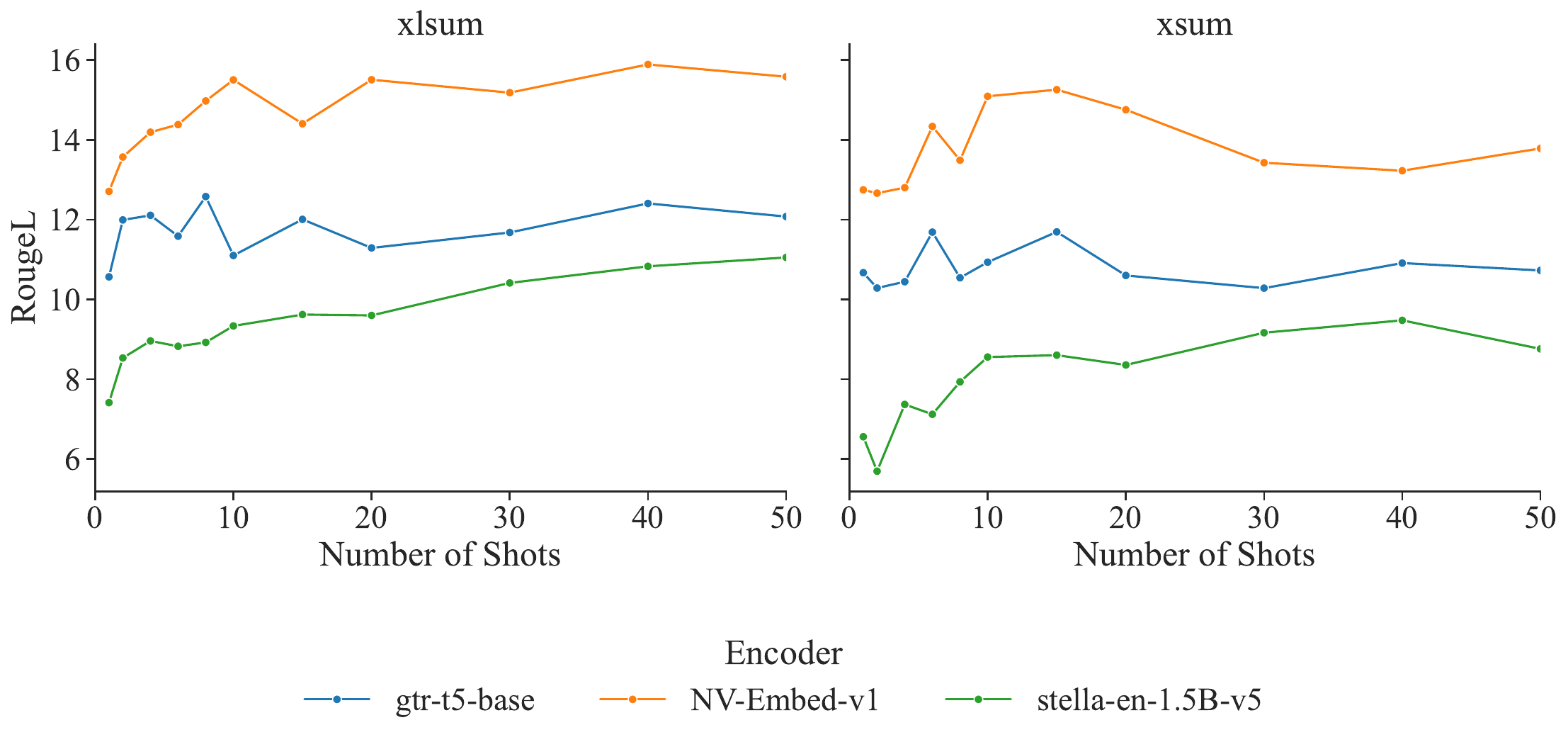}
    \caption{Text Summarization - Pretrained Projectors}
\end{figure}

%% file: tex/figures/text_summarization_finetuned.tex
\begin{figure}[ht]
    \centering
    \includegraphics[keepaspectratio, width=\textwidth]{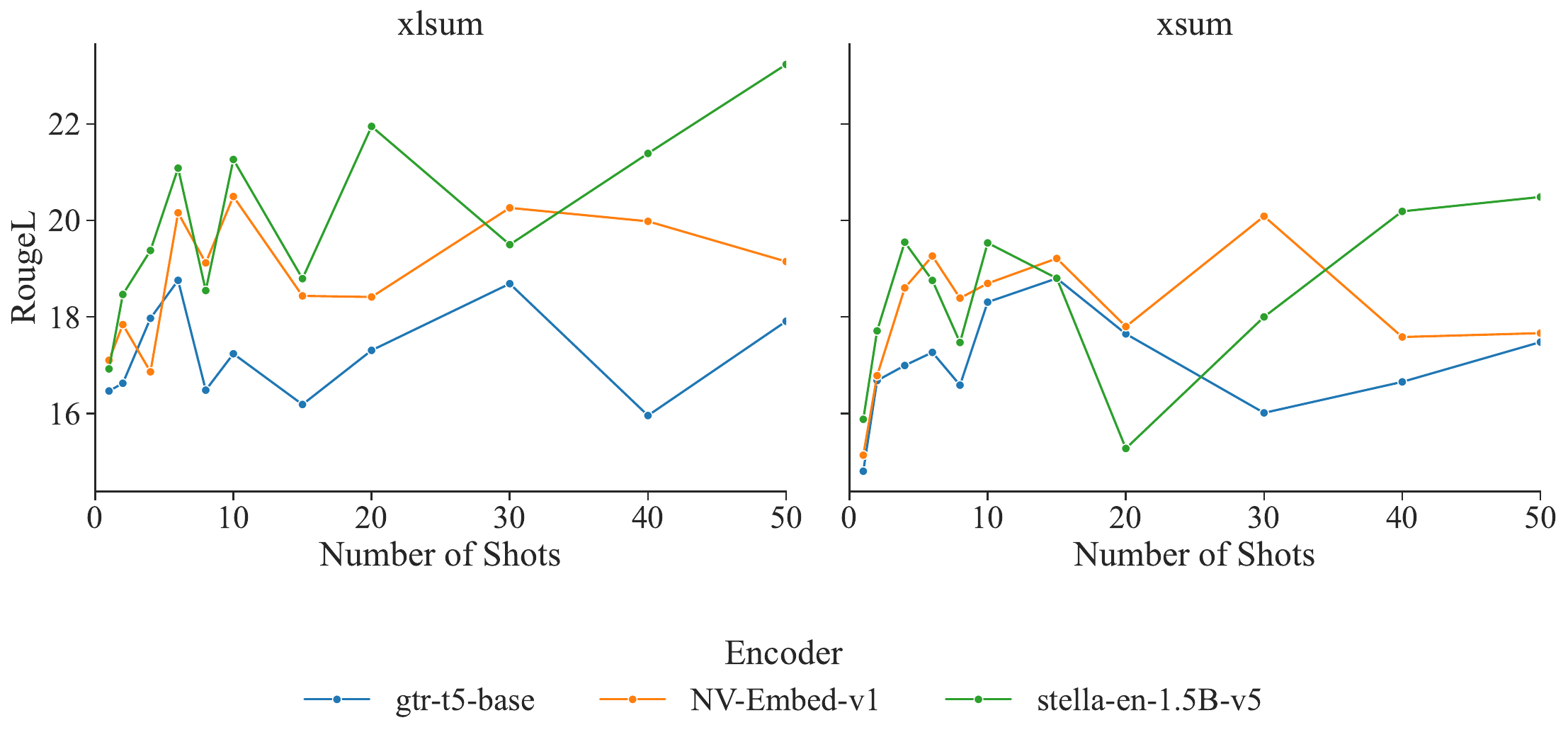}
    \caption{Text Summarization - Finetuned Projectors}
\end{figure}

%% file: tex/tables/training_config.tex
\begin{table}[t]
    \centering
    \caption{Hyperparameters for \method training.}
    % \resizebox{\linewidth}{!}{
    \small
    \begin{tabular}{l|l}
        \toprule
        \textbf{Hyperparameter} & \textbf{Value} \\
        \midrule
        Learning Rate & 1e-3 \\
        Learning Rate Schedule & Cosine Annealing \\
        Optimizer & AdamW \\
        $\beta_1$ & 0.9 \\
        $\beta_2$ & 0.999 \\
        Training dtype & bf16 \\
        Batch Size & 128 \\
        Generation Temperature & 2e-1 \\
        \bottomrule
    \end{tabular}
    % }
    \label{tab:hyperparameters}
\end{table}

%% file: tex/tables/brain_question_table.tex
\begin{table}[ht]
    \centering
    \caption{Question Categories and Their Associated Questions}
    \label{tab:question_categories}
    \tiny
    \begin{tabular}{l|l}
    \toprule
    \textbf{Category} & \textbf{Questions} \\
    \midrule
    \textbf{Sensory and Perceptual Descriptions} & 
    Does the input mention or describe a taste? \\
    & Does the input mention or describe a sound? \\
    & Does the sentence include a specific sound or auditory description? \\
    & Does the input mention or describe a visual experience? \\
    & Does the input mention or describe a texture? \\
    & Does the sentence describe a sensory experience? \\
    & Does the input mention anything related to color? \\
    & Does the input mention or describe a smell? \\
    & Does the input mention anything related to eyes? \\
    & Does the sentence describe a visual experience or scene? \\
    & Does the input describe a specific texture or sensation? \\
    & Does the sentence describe a specific sensation or feeling? \\
    \hline
    \textbf{Social, Emotional, and Interpersonal Content} & 
    Does the input mention anything related to arguing? \\
    & Does the input mention anything related to empathy? \\
    & Does the sentence involve a discussion about personal or social values? \\
    & Does the input discuss a societal issue or social justice topic? \\
    & Does the input mention or describe high emotional intensity? \\
    & Does the sentence describe a relationship between people? \\
    & Does the input mention or describe highly positive emotional valence? \\
    & Does the input mention or describe highly negative emotional valence? \\
    & Does the input mention anything related to conflict? \\
    & Does the sentence describe a personal or social interaction that leads to a change or revelation? \\
    & Does the sentence express a philosophical or existential query or observation? \\
    & Does the sentence involve an expression of personal values or beliefs? \\
    & Does the sentence express a sense of belonging or connection to a place or community? \\
    & Is the sentence emotionally positive? \\
    \hline
    \textbf{Cognitive and Reflective Aspects} & 
    Is the sentence reflective, involving self-analysis or introspection? \\
    & Does the input involve planning or organizing? \\
    & Does the text include a planning or decision-making process? \\
    & Does the sentence convey a decision or choice made by the narrator? \\
    & Does the sentence describe a personal reflection or thought? \\
    & Is the input about a discovery or realization? \\
    & Does the input contain a sense of ambiguity? \\
    & Is the sentence providing an explanation or rationale? \\
    & Does the input mention anything related to knowledge? \\
    \hline
    \textbf{Language, Structure, and Syntax} & 
    Does the sentence contain a proper noun? \\
    & Does the sentence include a conditional clause? \\
    & Does the sentence contain a negation? \\
    & Does the sentence use a unique or unusual word? \\
    & Does the sentence include a direct speech quotation? \\
    & Does the sentence include dialogue? \\
    & Does the sentence contain a cultural reference? \\
    & Does the sentence involve a recount of a social or community event? \\
    & Does the input include a comparison or metaphor? \\
    & Does the sentence include technical or specialized terminology? \\
    & Is the sentence abstract rather than concrete? \\
    & Does the sentence include an account of a miscommunication or misunderstanding? \\
    & Does the text describe a mode of communication? \\
    \hline
    \textbf{Physical Actions and Movements} & 
    Is the sentence conveying the narrator's physical movement or action in detail? \\
    & Does the input mention anything related to motor movements? \\
    & Does the sentence describe a physical action? \\
    & Does the sentence describe a physical sensation? \\
    & Does the sentence describe an activity related to daily life or routine? \\
    & Does the input mention anything related to an action? \\
    & Does the sentence involve spatial reasoning? \\
    % \hline
    % \textbf{Environment and Setting} & 
    % Is there mention of a city, country, or geographic feature? \\
    % & Does the sentence involve a description of physical environment or setting? \\
    % & Does the sentence mention a specific location? \\
    % & Does the sentence reference a specific time or date? \\
    % & Is time mentioned in the input? \\
    % & Does the input mention anything related to navigation? \\
    % & Does the text describe a journey? \\
    \hline
    \textbf{Numbers and Quantitative Information} & 
    Does the input mention a number less than 5? \\
    & Does the input contain a number? \\
    & Does the input mention a number greater than 100? \\
    & Does the input mention anything related to arithmetic? \\
    & Does the input mention anything related to calculation? \\
    & Does the input contain a measurement? \\
    \hline
    \textbf{Health, Disease, and Biological Aspects} & 
    Does the input mention anything related to diseases? \\
    & Does the input mention anything related to food? \\
    & Does the input mention anything related to age? \\
    & Does the input mention anything related to gender? \\
    & Does the input mention anything related to disgust? \\
    & Does the input mention anything related to children? \\
    \hline
    \textbf{Conflict, Urgency, and Change} & 
    Does the sentence involve an unexpected incident or accident? \\
    & Does the input mention anything related to anger? \\
    & Does the sentence convey a sense of urgency or haste? \\
    & Does the sentence describe a change in a physical or emotional state? \\
    & Does the sentence describe a moment of relief or resolution of tension? \\
    & Does the sentence express the narrator's opinion or judgment about an event or character? \\
    % \hline
    % \textbf{Cultural, Historical Contexts} & 
    % Does the text include a reference to a past era or time period? \\
    % & Is the input related to a specific industry or profession? \\
    % \hline
    % \textbf{Objects, Things} & 
    % Does the sentence involve the mention of a specific object or item? \\
    % & Does the story involve a personal project or creation? \\
    % & Does the sentence include a personal anecdote or story? \\
    \bottomrule
    \end{tabular}
\end{table}